\pdfoutput=1

\documentclass[11pt]{article}

\usepackage[]{acl}

\usepackage{url}
\usepackage{times}
\usepackage{latexsym}
\usepackage{amsmath} 
\usepackage{amsfonts} 
\usepackage{multirow}
\usepackage{color, colortbl} 
\usepackage{subcaption} 

\usepackage[T1]{fontenc}
\usepackage{graphicx}
\usepackage{booktabs}
\usepackage{float} 

\usepackage[utf8]{inputenc}

\usepackage{microtype}

\usepackage{inconsolata}

\newcommand\blfootnote[1]{%
  \begingroup
  \renewcommand\thefootnote{}\footnote{#1}%
  \addtocounter{footnote}{-1}%
  \endgroup
}

\title{Investigating the Emergent Audio Classification Ability of ASR Foundation Models}

\author{Rao Ma$^\ast$, Adian Liusie$^\ast$, Mark J. F. Gales, Kate M. Knill \\
  ALTA Institute, Department of Engineering, University of Cambridge \\
  \texttt{rm2114@cam.ac.uk, al826@cam.ac.uk, mjfg@eng.cam.ac.uk, kmk1001@cam.ac.uk} \\}

\definecolor{Gray}{gray}{0.9}

\begin{document}
\maketitle
\begin{abstract}
Text and vision foundation models can perform many tasks in a zero-shot setting, a desirable property that enables these systems to be applied in general and low-resource settings. There has been far less work, however, on the zero-shot abilities of ASR foundation models, with these systems typically fine-tuned to specific tasks or constrained to applications that match their training criterion and data annotation. In this work we investigate the ability of Whisper and MMS, ASR foundation models trained primarily for speech recognition, to perform zero-shot audio classification. We use simple template-based text prompts at the decoder and use the resulting decoding probabilities to generate zero-shot predictions. Without training the model on extra data or adding any new parameters, we demonstrate that Whisper shows promising zero-shot classification performance on a range of 8 audio-classification datasets, outperforming the accuracy of existing state-of-the-art zero-shot baselines by an average of 9\%. One important step to unlock the emergent ability is debiasing, where a simple unsupervised reweighting method of the class probabilities yields consistent significant performance gains. We further show that performance increases with model size, implying that as ASR foundation models scale up, they may exhibit improved zero-shot performance. 
\end{abstract}

\blfootnote{$^\ast$ Equal Contribution.}

\section{Introduction}
The evolution of large-scale pre-trained foundation models has reshaped the way various complex tasks are approached. Large language models (LLMs) have been trained over massive text corpora \cite{radford2019language, brown2020language, chung2022scaling, touvron2023llama} and can be used out of the box for diverse NLP tasks. Similarly, vision-to-text models, such as those trained to predict image captions, have facilitated zero-shot transferability for image classification \cite{li2017learning, radford2021learning}. A fascinating property of these systems is their emergent abilities, where the systems can be applied effectively to a wide range of tasks that were not seen during training \cite{bang2023multitask}. This capability can serve as an alternative to task-specific approaches or further fine-tuning.


\begin{figure}[t]
    \centering
    \includegraphics[width=\linewidth, trim={5mm 0 4mm 0}]{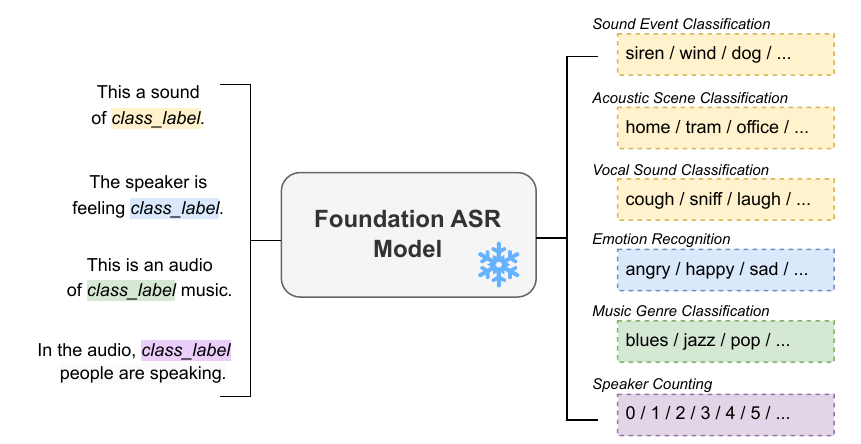}
    \caption{This paper looks at zero-shot prompting of ASR foundation models for audio classification, without any further training or introducing any new parameters. We use task-specific prompts and evaluate on various downstream tasks and datasets.}
    \label{fig:high_level}
\end{figure}

Despite the progress in text and vision models, there has been limited work done to investigate the general zero-shot ability of speech-based models. \citet{peng2023prompting} recently demonstrated that Whisper can be prompted for zero-shot task generalization, however their focus is on three forms of speech recognition tasks, and therefore remains close to the original pre-training task domain. Further, \citet{elizalde2023clap} use contrastive pre-training to match representations from audio and text encoders, which can then be used to classify audio samples. The Contrastive Language-Audio Pretraining (CLAP) approach, however, was trained in a fashion that matched its downstream evaluation tasks, and the further the task domain diverged from the training domain, the worse the task transferability.

This work investigates the abilities of Automatic Speech Recognition (ASR) systems when applied to tasks that they were not explicitly trained on during training. It focuses on task transferability and examines whether speech foundation models such as Whisper \cite{radford2023robust} and MMS \cite{pratap2023scaling} demonstrate any zero-shot task transferability, with a particular focus on zero-shot audio classification. We demonstrate that without updating or adding any parameters, Whisper can be prompted to achieve state-of-the-art zero-shot performance for downstream audio classification tasks. 8 data sets from 6 downstream tasks are used for evaluation (Figure \ref{fig:high_level}) and we show that Whisper performs significantly better than random, and on average 9.2\% higher than the CLAP baseline~\cite{elizalde2023clap}. Further, our work highlights the importance of task calibration for unlocking the zero-shot capabilities, where unsupervised reweighting of the probabilities yields performance improvements of up to 25\%. We perform ablations on prompts, model family and model size to analyze the observed phenomenon and test the generalizability of our proposed zero-shot prompting methodology. Further, we provide a preliminary investigation of Whisper on audio question answering and demonstrate that Whisper can be prompted to answer questions on input audio in a zero-shot fashion, with performance significantly better than random.  

\section{Related Work}
\textbf{Emergent Abilities of LLMs} \citet{wei2022emergent} demonstrate that LLMs gain emergent abilities where certain task abilities emerge sharply at certain model sizes, however, \citet{schaeffer2023emergent} present a contrasting perspective and question whether these observations are caused by the choice of evaluation metric. Nonetheless, it has been demonstrated that if scaled sufficiently, LLMs can gain impressive abilities that the model was never explicitly trained for. Examples include in-context few-shot learning ability \cite{brown2020language}, zero-shot task transfer \cite{radford2019language}, and zero-shot reasoning abilities \cite{kojima2022large}. In this work we refer to emergence as {\em when a model acquires an ability that the model wasn't explicitly trained to achieve}, and consider similar emergent zero-shot task transfer of audio models.


\vspace{1.5mm}
\noindent \textbf{Prompting LLMs} Early forms of prompting employed simple keyword-based inputs or fill-in-the-blank style prompts \cite{schick-schutze-2021-exploiting, gao-etal-2021-making}, where impressive few-shot performance was observed by framing new tasks within the format of the pre-training task. For generative transformers,  prompting was extended by using natural language prompts to differentiate between tasks \cite{radford2019language, sanh2021multitask} or for providing few-shot examples \cite{brown2020language}. Further developments in the field introduced additional training stages, such as instruction-tuning \cite{ouyang2022training} and supervised fine-tuning \cite{chung2022scaling}, to enhance model alignment and enable better instruction-following abilities of models for zero-shot task completion. 

\vspace{1.5mm}
\noindent \textbf{Debiasing Zero-Shot Decisions} 
GPT-3 classification decisions were shown to be sensitive to factors such as the ordering of examples and choice of label words. \citet{zhao2021calibrate} demonstrated that a context-dependent `null input' could be used to debias decisions, which yields substantial performance gains. Similarly, \citet{liusie2023mitigating} demonstrated that one can apply prior-matching to yield globally all-calibrated predictions which improves zero-shot classification robustness. Debiasing can also be done through prompt design; \citet{guo-etal-2022-auto} search for cloze-style prompts that have stereotypical biases, and fine-tune the models to minimize disagreement. 

\vspace{1.5mm}
\noindent \textbf{Adapting ASR Foundation Models} ASR Foundation models have been adapted to downstream tasks through fine-tuning, such as for disfluency removal and spoken grammatical error correction \cite{bannò2023endtoend}, or as an E2E spoken language understanding system \cite{wang2023whislu}. Further, \citet{gong_whisperat} freeze Whisper and train a lightweight audio tagging model, and demonstrate good performance for downstream audio classification tasks. \citet{wang2023can} shows that test-time adaptation of Whisper for Chinese dialect ASR can be achieved with speech-based in-context learning. Lastly, \citet{elizalde2023clap} use contrastive pre-training to match representations from audio and text encoders, and fine-tune the representations for downstream audio classification tasks. 


\begin{figure*}[t]
    \centering
    \includegraphics[width=\linewidth]{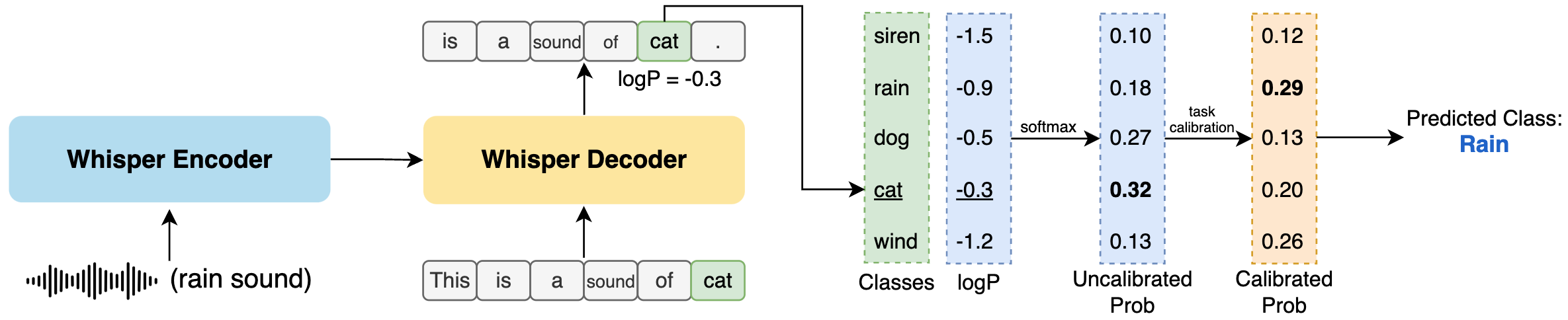}
    \caption{ASR foundation models are leveraged for zero-shot audio classification by prompting the decoder to calculate the log-likelihood of label sequences associated with each class. The log-likelihood for each class is converted to probabilities and post-processed to a predicted class. This process is displayed for Whisper.}
    \label{fig:overall_approach}
\end{figure*}

\section{Zero-Shot Classification of ASR Foundation Models}
This paper investigates the emergent zero-shot audio classification abilities of large-scale ASR foundation models. These systems are trained specifically for speech recognition and were not explicitly trained for any of the downstream classification tasks considered in this paper. We question whether one can use prompting to leverage the implicit knowledge learned from pre-training to achieve various audio classification tasks.

\subsection{Zero-shot Prompting}
\label{sec:default}
In this work, we use a simple template filling prompting strategy, where given an input audio sample, we assess the probability of decoding a label sequence associated with each classification class (as shown in Figure \ref{fig:overall_approach}). We leverage various `prompts' by considering different templates to represent the label sequences (as shown in Figure \ref{fig:high_level}). To convert likelihoods to class probabilities, we treat the ASR system as a generative classifier:

Let $P_{\theta}(x|s)$ represent the likelihood associated with ASR decoding the word sequence $x \! \in \!\mathcal{X}$ given an input audio $s$. Let $y \in \{ \omega_1,  \omega_2, ...  \omega_K \}$ be one of $K$ possible output classes, and $t(\omega_k) \! \in \! \mathcal{X}$ represent a particular mapping of a class to a word sequence representing the class. We assume that the zero-shot ASR classification probability $\tilde{P}_{\theta}$ for a particular class is proportional to the likelihood of generating each respective class label sequence given the input audio: 
\begin{equation}
    \tilde{P}_{\theta}(y=\omega_k| s) = \frac{P_{\theta}(t(\omega_k)| s)}{\sum_{\omega_j} P_{\theta}(t(\omega_j)| s)}
\end{equation}

\noindent The model's prediction is then the class with the highest associated probability:
\begin{equation}
    \hat{y} = \underset{\omega}{\text{argmax}} \; \tilde{P}_{\theta}(\omega| s)
\end{equation}

\subsection{Task Calibration}
A concern with the zero-shot prompting approach described above is the potential presence of implicit biases. Previous works have demonstrated that zero-shot generative classifiers may have associated biases that can degrade performance \cite{zhao2021calibrate, liusie2023mitigating}. For example, the model may favour words that are common in pre-training, which may lead to predictions being skewed towards particular classes.


To account for misaligned model probabilities, approaches exist to modify model outputs to be better aligned of which the most prominent example is model calibration. The objective of model calibration is for the top-1 confidences to better reflect the expected accuracy of decisions:
\begin{equation}
    \frac{1}{N} \sum_{i=1}^N P_{\theta}(\hat{y}^{(i)}|s^{(i)}) = \frac{1}{N} \sum_{i=1}^N \delta(\hat{y}^{(i)}\!=\!y^{(i)})
\end{equation}
where $y^{(i)}$ is the reference classification label for audio $s^{(i)}$.
Model calibration (sometimes referred to as top-label calibration) is typically performed in a post-hoc fashion \cite{isotonic1972, platt1999probabilistic, guo2017calibration}, where it is often assumed that the ordering of the classes is valid and so a monotonic function can be applied to scale probabilities, without altering the ordering. Since these standard model calibration approaches do not change the output prediction order, however, they will be ineffective in cases where the model demonstrates systematic class biases, as the system will remain biased towards particular classes.


To address this concern, a different calibration approach is required that can change the ordering of decisions and the top-1 decision. We refer to such an approach as \textbf{task calibration}, since such calibration may be most necessary when there is a mismatch between the training and downstream task. For task calibration, the system should be altered to provide global all-label calibrated decisions, such that for each class, the system confidence accurately represents the expected accuracy. 
\begin{equation}
    \fontsize{10.5}{10}\selectfont
    \frac{1}{N} \! \sum_{i=1}^N \! P_{\theta}(\omega_k|s^{(i)}) \!=\! \frac{1}{N} \! \sum_{i=1}^N \! \delta(y^{(i)}\!=\!\omega_k) \quad \forall \omega_k
    \label{eq:all-label-calibration}
\end{equation}

\noindent Note that all-label global calibration is not a sufficient condition, and may have limitations. To illustrate this, if the labels have a uniform true prior, the only valid solution with temperature annealing is the trivial solution of infinite temperature which yields random performance. Therefore one has to select approaches that sensibly debias the model, and in this work, two particular forms of task calibration are considered.





\subsubsection{Prior Matching}
\label{sec:prior_match}
The first task calibration method we consider applies global all-label calibration. Following \citet{liusie2023mitigating}, one can reweight the outputs of the classifier by introducing weights $\alpha_{1:K}$ to rescale the probabilities. 

\begin{equation}
    \hat{P}_{\theta}(\omega_k| s, \alpha_{1:K}) = \frac{\alpha_k \tilde{P}_{\theta}(\omega_k| s)}{\sum_j \alpha_j \tilde{P}_{\theta}(\omega_j| s)}
\end{equation}

\noindent Assuming that unsupervised data is available for a particular task (or if all the evaluation is available as an unsupervised set), the output probabilities can be reweighted to ensure that the corresponding output prior matches the expected true prior, done by finding the weights $\bar{\alpha}_{1:K}$ that ensure such a prior,
\begin{equation}
    \hat{P}_{\theta}(\omega_k| \alpha_{1:K}) = \mathbb{E}_{s} \{ \hat{P}_{\theta}(\omega_k|s, \alpha_{1:K})\ \}
    \label{eq:model_calibration}
\end{equation}
\begin{equation}
    \bar{\alpha}_{1:K} = \underset{\alpha_{1:K}}{\text{argmin}} \sum_{\forall \omega} | \hat{P}_{\theta}(\omega| \alpha_{1:K}) - P(\omega)| 
    \label{eq:task_calibration}
\end{equation}

\noindent Where $P(\omega)$ is the true prior for the considered task. In cases where the underlying class distribution is not known, the prior can be assumed to be uniform, $P(\omega)\!=\!\frac{1}{K}$, which is the assumption made throughout this paper. The solution has a single free variable, but by constraining $\alpha_1 \! = \!1$ one can find an exact solution that perfectly matches the prior, a search which can be done efficiently. Note that such a solution (equation \ref{eq:task_calibration}) satisfies global all-label calibration (equation \ref{eq:all-label-calibration}), but not necessarily top-1 model-calibration (equation \ref{eq:model_calibration}). 

\subsubsection{Null-Input Calibration} 
\label{sec:null_input}
The previous method requires unsupervised data, which in some settings can be a drawback. \citet{zhao2021calibrate} proposed a data-free method which uses a null-input, $\phi$, to estimate the weights, which \citet{liusie2023mitigating} demonstrate is an approximation of prior-matching, 
\begin{equation}
    \bar{\alpha}_{k}\approx \frac{1}{\mathbb{E}_{s} \{{P}_{\theta}(\omega_k| s)\}} \approx \frac{1}{P_\theta (\omega_k|\phi)}
\end{equation}

\noindent i.e. the null input is used as the audio input $s$, and with prompting one can get an output probability distribution. This may be indicative of bias since the null-input should yield a uniform 
pmf output, and this is used to correct all downstream decisions. 

For LLMs, the null-input $\phi$ is designed to be an input with no information, e.g. an empty string or the input `N/A'. For our work, using text-based null-inputs is not applicable. Therefore, for speech recognition models, we propose using two different forms of null-inputs: using a sequence of all zero vectors as the input of the encoder, or using acoustic features generated from synthetic Gaussian white noise with $\sigma=1$. 


\section{Experimental Set Up}
\subsection{Models}
Two ASR foundation models are considered: Whisper \cite{radford2023robust} and the Massively Multilingual Speech (MMS) model \cite{pratap2023scaling}.

\vspace{1.5mm}
\noindent\textbf{Whisper} \cite{radford2023robust} is an encoder-decoder transformer model trained on 680K hours of labelled speech data obtained through large-scale weak supervision. Whisper checkpoints come in varying sizes, ranging from 39M parameters (Whisper tiny) to 1.55B parameters (Whisper large), available either as English-only or multilingual models. The largest model is only available in the multilingual version. Whisper is trained for automatic speech recognition and voice activity detection, with the multilingual models further trained for speech translation and language identification. 

\vspace{1.5mm}
\noindent\textbf{MMS} \cite{pratap2023scaling} is a CTC model which has a decoder that is a simple linear layer mapping to a set of characters. The model has 1B parameters and is first pre-trained on 491K hours of unlabelled data using self-supervised training. For multilingual speech recognition, the model is further trained on 45K hours of labelled data spanning 1,107 languages, data collected by aligning New Testament audios and texts. 

\subsection{Datasets}
We assess our systems across 8 diverse audio classification datasets, encompassing 6 distinct tasks. Sound Event Classification (SEC) comprises of \textbf{ESC50} \cite{piczak2015esc} (50 environmental sounds) and \textbf{UrbanSound8K} \cite{salamon2014dataset} (10 urban sounds). Acoustic Scene Classification (ASC) uses \textbf{TUT2017} \cite{mesaros2016tut}, featuring 15 acoustic scenes spanning both outdoor and indoor environments. Vocal Sound Classification (VSC) uses \textbf{Vocal Sound} \cite{gong2022vocalsound} with 6 distinct human vocal sound categories. Emotion Recognition (ER) comprises of \textbf{RAVDESS} \cite{luna2021proposal} and \textbf{CREMA-D} \cite{cao2014crema}, each containing speakers expressing 8 and 6 different emotions, respectively. Music Genre Classification (MGC) uses \textbf{GTZAN} \cite{sturm2013gtzan}, containing music classified into 10 genres. Additionally, Speaker Counting (SC) uses \textbf{LibriCount} \cite{stoter2018libricount}, featuring audio clips with varying speaker counts from 0 to 10. Complete dataset statistics are outlined in Table \ref{tab:dataset}. Five of the datasets are balanced over the classes (ESC50, TUT2017, Vocal, GTZAN, and LibriCount), while the other three have slightly imbalanced distributions. 

\begin{table}[H]
    \centering
    \footnotesize
    \begin{tabular}{ll|ccc}
    \toprule
       Task & Dataset & Utts & Avg. Dur. & $K$ \\
        \midrule
        \multirow{2}*{SEC} & ESC50 & 2,000 & 5.0 & 50 \\
        & UrbanSound8K & 8,732 & 3.6 & 10 \\
        \midrule
        ASC & TUT2017 & 1,620 & 10.0 & 15 \\
        \midrule
        VSC & Vocal Sound & 3,594 & 5.0 & 6 \\
        \midrule
        \multirow{2}*{ER} & RAVDESS & 1,440 & 3.7 & 8 \\
        & CREMA-D & 7,442 & 5.0 & 6 \\
        \midrule
        MGC & GTZAN & 1,000 & 30.0 & 10 \\
        \midrule
        SC & LibriCount & 5,720 & 5.0 & 11 \\
    \bottomrule
    \end{tabular}
    \caption{Test set statistics, displaying the total number of test utterances, the average duration of each audio sample (in seconds), and the number of classes $K$.}
    \label{tab:dataset}
\end{table}


\subsection{Method}
\begin{table}[H]
    \small
    \centering
    \begin{tabular}{l|l}
    \toprule
    Task & \multicolumn{1}{c}{Prompt} \\
    \midrule
    ER & The speaker is feeling \textit{class\_label}. \\
    MGC & This is an audio of \textit{class\_label} music. \\
    SC & In the audio, \textit{class\_label} people are speaking. \\
    others & This is a sound of \textit{class\_label}. \\
    \bottomrule
    \end{tabular}
    \caption{Manually designed prompts used for each task. The bottom prompt is used for SEC, ASC and VSC.}
    \label{tab:prompts}
\end{table}

The default prompts used for the different tasks are shown in Table \ref{tab:prompts}, which were adapted from the prompts of \citet{elizalde2023clap}. We calculate class probabilities using our three methods\footnote{The code is made available at \url{https://github.com/JuliRao/Whisper_audio_classification}.}; the base `uncalibrated' probabilities, prior matching, and the null-input strategy (both zero-inputs and Gaussian white noise). For the Gaussian white noise null-input, the $\sigma=1$ and the synthetic clips are generated to have the same average duration as the task's clips. When employing Whisper as the underlying speech model, we calculate the probability that the decoder generates each prompted text sequence with teaching-forcing. For MMS, we apply a dynamic programming algorithm \cite{graves2006connectionist} to compute the probability of generating a sentence for the given input audio. All the possible alignments are considered in this process. 

\subsection{Baselines}
\label{sec:baselines}
We compare our performance against AudioCLIP \cite{guzhov2022audioclip} and CLAP \cite{elizalde2023clap}. CLIP \cite{radford2021learning} is a multimodal system that generates representations for images and text, which AudioCLIP extends to also incorporate the audio modality. They introduce an audio head and perform contrastive learning on AudioSet (a sound event classification dataset) to align the audio embeddings with the other modalities. CLAP adopts a similar approach and aligns a pre-trained text encoder with a pre-trained audio encoder using contrastive learning. The model is trained using a sound event classification dataset and three audio captioning datasets. In CLAP, the text encoder uses target sequences written as natural language sentences rather than single-class words.

\subsection{Supervised Baseline}
\label{sec:baseline_details}
To consider the performance gap between zero-shot Whisper and supervised approaches, we further consider fine-tuning Whisper on training data to obtain an upper bound of supervised model performance. This is done on TUT and Vocal, which have available training data sets. We perform supervised training with parameter efficient fine-tuning approaches; LoRA \cite{hu2021lora} and soft prompt tuning (SPT) \cite{lester2021power, ma2023adapting}. During training, the audio clip is provided to the model encoder and the model decoder is trained to generate the corresponding class label. 

We note here that unsurprisingly, the zero-shot performance was considerably worse than the supervised fine-tuning results. Therefore, although the results section will demonstrate that Whisper can show impressive zero-shot task transfer to unseen audio classification tasks, in settings where labelled data is available, fine-tuning will yield better performance. More details on the supervised training details and experimental results can be found in Appendix \ref{sec:appendix_sup}.

\begin{table*}[t]
  \centering \footnotesize
   \renewcommand\tabcolsep{5pt}
\begin{tabular}{l|c|c|c|c|c|c|c|c|c}
        \toprule
        Model & ESC50 & US8K  & TUT & Vocal & RAVDESS & CREMA-D & GTZAN & LibriCount & \textbf{Avg.} \\
        \midrule
        \rowcolor{Gray}
        \multicolumn{5}{c}{Baselines (\S \ref{sec:baselines})} & \multicolumn{5}{l}{}  \\        
        Random & 2.0 & 10.0 & 6.7 & 16.7 & 12.5 & 16.7 & 10.0 & 9.1 & 10.4 \\
        AudioCLIP & 69.4 & 65.3 & - & - & - & - & - & - & - \\
        CLAP  & 82.6 & 73.2 & 29.6 & 49.4 & 16.0 & 17.8 & 25.2 & 17.9 & 39.0 \\
        \rowcolor{Gray}
        \multicolumn{5}{c}{Uncalibrated (\S \ref{sec:default})} & \multicolumn{5}{l}{}  \\        
        MMS large (1B)           & 1.7  & 9.6  & 4.9 & 14.2 & 13.5 & 17.2 & 8.3  & 8.4 & 9.7  \\
        Whisper medium.en (769M) & 27.9 & 39.5 & 7.2 & 59.0 & 15.3 & 20.9 & 15.2 & 8.2 & 24.2 \\ 
        Whisper medium (769M)    & 29.7 & 45.8 & 7.5 & 44.6 & 16.7 & 19.9 & 28.4 & 9.4 & 25.2 \\ 
        Whisper large-v2 (1.6B)     & 38.9 & 50.5 & 7.7 & 60.1 & 15.1 & 20.2 & 38.2 & 9.2 & 30.0 \\
        \rowcolor{Gray}
        \multicolumn{5}{c}{Prior-matched (\S \ref{sec:prior_match})} & \multicolumn{5}{l}{}  \\        
        MMS large (1B)           & 2.4  & 10.9 & 7.6 & 11.5 & 12.2 & 17.2 & 10.5 & 11.5 & 10.5  \\
        Whisper medium.en (769M) & 56.2 & 60.9 & 18.3 & 82.8 & 29.0 & 22.6 & 29.7 & 9.8  & 38.7 \\ 
        Whisper medium (769M)    & 57.5 & 61.6 & 25.2 & 82.4 & 35.0 & 25.9 & 48.6 & 16.3 & 44.1 \\ 
        Whisper large-v2 (1.6B)     & 65.4 & 60.4 & 26.0 & 84.9 & 41.7 & 28.8 & 60.9 & 17.3 & \textbf{48.2} \\
        \bottomrule
    \end{tabular}
    \vspace{-4pt}
  \caption{Baseline and zero-shot task performance using the default prompts (of Table \ref{tab:prompts}). The classification accuracy for each individual task, as well as the average accuracy across all eight tasks, is reported.}
  \label{tab:main_table}
\end{table*}

\subsection{Evaluation} The focus of this work is on zero-shot classification performance and therefore the top-1 accuracy of the test data is used as the main performance metric for all systems. For Whisper and MMS, the test utterances are down-sampled to 16kHz to match the pre-training procedure. CLAP uses a higher sampling rate of 44.1kHz in the audio encoder, which is more computationally expensive. 


\section{Results}
\subsection{Audio Classification Performance}

Table \ref{tab:main_table} shows the audio classification results for 3 Whisper systems and 1 MMS system for our 8 datasets, with comparisons to random performance and relevant baselines. We display our zero-shot prompted performance when using either base output ASR likelihoods (\S \ref{sec:default}) and when post-processing the outputs using prior-matching (\S \ref{sec:prior_match}). We observe the following points:  

\noindent \textbf{1) Whisper performs zero-shot audio classification better than random.} Using simple template prompts and output likelihoods, Whisper large-v2 achieves an average zero-shot accuracy of 30\%, considerably better than the average random performance. Further, increasing parameter size yields a performance boost (769M to 1.6B parameters) and the multilingual Whisper performs better than the English-only model for the medium size. 


\noindent \textbf{2) MMS fails for zero-shot audio classification.} This could be explained by the fact that MMS is forced to generate either an output token or a blank symbol for every input frame. 
The text prompt, including the classification symbols, must be aligned with the mismatched input audio.
Using the resulting probability for classification may prove challenging when generalizing to zero-shot audio classification tasks. For Whisper, the attention mechanism allows it to attend over the entire input sequence to capture high-level audio information.

\noindent \textbf{3) Prior Matching yields large performance improvements.} By reweighting the output probabilities in an unsupervised fashion (i.e. without using the test labels), large performance boosts are observed for all Whisper systems. Whisper can now demonstrate reasonable performance for all 8 tasks, and reducing the inherent class bias leads to an improvement of average accuracy to 48.2\%. 

\noindent \textbf{4) Zero-Shot Whisper outperforms baselines, demonstrating our approach is a powerful zero-shot audio classification method.} Note that CLAP is tuned on sound event classification and audio captioning datasets, and has therefore been trained to be aligned with tasks such as ESC50 and US8K. Nonetheless, even including performance on these tasks, our approach outperforms CLAP by an average of 9.2\%, and has consistent and substantial performance improvements for most out-of-domain tasks.

\subsection{Robustness to Prompts}
Table \ref{tab:RAVDESS} displays RAVDESS performance for different prompts, with Whisper large-v2 and prior-matching. The first prompt is the default prompt used for the main experiments, prompts 2-4 contain only the class label, and prompts 5-9 were generated by asking ChatGPT to paraphrase\footnote{using the prompt: \textit{``Please paraphrase the given prompt five times with simple language:"}} prompt 1. The results show that, though zero-shot prompting can work for various prompts, there is considerable prompt sensitivity. Interestingly, although prompts 2-4 are closest to the pre-training task of ASR decoding, we observe that, on average, the natural language prompts demonstrate considerably better performance, implying that the zero-shot ability can be attributed to more than ASR task transfer. Further, ensembling all 9 prompts leads to the best performance of 44.0, a performance boost which was also observed for other tasks, as displayed in Table \ref{tab:prompt_ensemble}. Complete results for varying prompts for all datasets can be found in Appendix \ref{sec:appendix_prompts}.

\begin{table}[t]
    \centering
    \footnotesize
    \renewcommand\tabcolsep{3pt}
    \begin{tabular}{l|c}
    \toprule
    \multicolumn{1}{c|}{Prompt} & Acc \\
    \midrule
    The speaker is feeling \textit{class\_label}. & 41.7 \\
    \midrule
    \textit{class\_label} & 20.7 \\
    \textit{(class\_label)} & 33.1 \\
    \textit{[class\_label]} & 32.6 \\
    \midrule
    The person talking feels \textit{class\_label}. & 38.5 \\
    The speaker is experiencing \textit{class\_label} emotions. & 20.8 \\
    The person speaking is in a \textit{class\_label} mood. & 29.9 \\
    The speaker's emotion is \textit{class\_label}. & 33.6 \\
    The person talking is filled with \textit{class\_label} feelings. & 39.7 \\
    \midrule
    Ensemble of Prompts & 44.0 \\
    \bottomrule
    \end{tabular}
    \caption{Performance of Whisper large-v2 with different prompts on RAVDESS (using prior-matched outputs).}
    \label{tab:RAVDESS}
\end{table}

\begin{table}[!htbp]
\small
    \centering
    \begin{tabular}{l|c|c}
    \toprule
        Dataset & Default & Ensemble \\
    \midrule
        ESC50 & 65.4 & 67.1 \\
        US8K & 60.4 & 67.6 \\
        TUT & 26.0 & 25.2 \\
        Vocal & 84.9 & 87.3 \\
        RAVDESS & 41.7 & 44.0 \\
        CREMA-D & 28.8 & 33.1 \\
        GTZAN & 60.9 & 60.0 \\
        LibriCount & 17.3 & 22.0 \\
    \midrule
        Average & 48.2 & \textbf{50.8} \\
    \bottomrule
    \end{tabular}
    \caption{Performance of the default prompt and the ensemble of 9 prompts on audio classification tasks.}
    \label{tab:prompt_ensemble}
\end{table}


\subsection{Null-input Performance}
Prior matching requires a set of unlabelled test data, and is not applicable when a single/few samples have to be classified. In such settings, the null-input approximation (\S \ref{sec:null_input}) can be used as a zero-resource debiasing approach, which can use either all-zeros in the encoder input or Gaussian noise. Table \ref{null-input} demonstrates that, compared to the uncalibrated baseline results, null-input debiasing improves model performance by an average of 6.7\% and 4.8\% over all models and tasks for the 2 methods respectively. These results show that the null-input method can provide a performance boost via data-free calibration, however, there is still a considerable gap with prior-matching performance. More detailed results can be found in Appendix \ref{sec:appendix_main}.

\begin{table}[H]
\centering 
\footnotesize
\begin{tabular}{l|c|c|c}
    \toprule
    Method & medium.en & medium & large-v2 \\
    \midrule
        Uncalibrated & 24.2 & 25.2 & 30.0 \\
        \midrule
        Zero Input & 29.8 & 34.8 & 34.9 \\
        Gaussian Noise & 28.5 & 29.5 & 35.8 \\
        \bottomrule
    \end{tabular}
  \caption{Average accuracy of 8 audio classification tasks with null-input calibration.}
  \label{null-input}
\end{table}


\subsection{Analysis of Predicted Distribution}
\begin{figure}[H]
    \centering
    \includegraphics[width=\columnwidth]{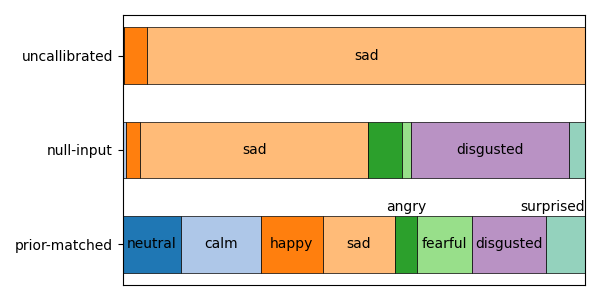}
    \caption{Predicted class distribution for Whisper large-v2 on RAVDESS. Bar width is proportional to the fraction of decisions per class.}
    \label{fig:distribution}
\end{figure}

To analyze the performance boost observed from debiasing, Figure \ref{fig:distribution} illustrates the output class distributions on RAVDESS for the various methods. We observe that the uncalibrated outputs are strongly dominated by the `sad' class. Using the null-input method (where we select to use the zero-input approach) still yields relatively imbalanced decisions. However, we observe that prior-matching (by design) leads to a more balanced distribution of predictions. Equivalent plots are shown for different datasets in Appendix \ref{sec:appendix_pred}.

\subsection{Ability with Scale}
\begin{figure}[H]
    \centering
    \includegraphics[width=\columnwidth]{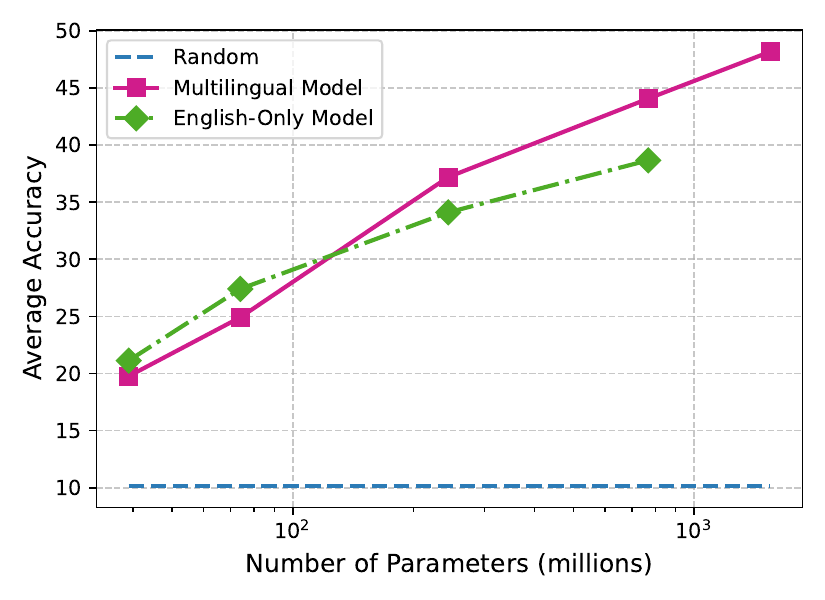}
    \caption{Parameter size vs average accuracy (with prior-matching) for different versions of Whisper models.}
    \label{fig:model_size}
\end{figure}

Figure \ref{fig:model_size} illustrates the improvement of average ability over all tasks as the model size increases. We observe a continuous improvement in performance as the model size increases, and secondly beyond 500M parameters the multilingual models achieve much better performance than the English-only models (when comparing models of similar size). This may be due to the increased training data, as well as the multi-task pre-training criterion (which includes speech translation and language identification as well).  

\subsection{Audio Question Answering}
The previous experiments demonstrated that Whisper can be zero-shot prompted to perform a multitude of audio classification tasks with reasonable performance. Here, we provide an initial investigation into the ability of Whisper for the more challenging task of audio question answering. 

\begin{figure}[h]
    \centering
    \includegraphics[width=\linewidth]{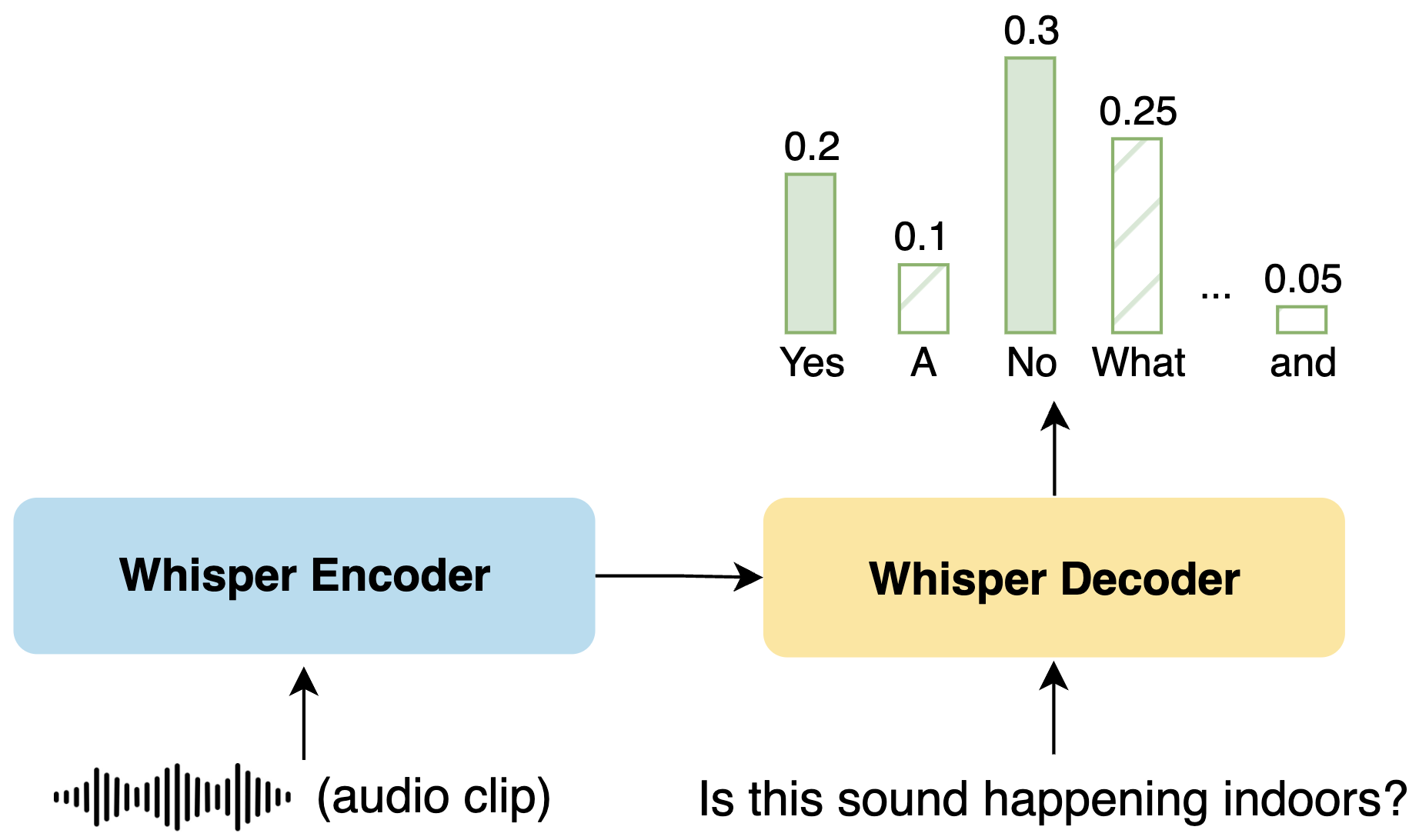}
    \caption{Zero-shot audio question answering method.}
    \label{fig:aqa_method}
\end{figure}

Clotho-AQA \cite{lipping2022clotho} is a dataset of audio clips selected from the Clotho dataset, with corresponding questions and answers collected through crowd-sourcing. Our experiments focus on the yes-no questions of Clotho-AQA, where each question is a yes-no question corresponding to an input audio sample, with three independent `yes' or `no' annotations. We consider both the `majority' set, where the label is assigned as the most select options, and `unanimous' set, where the questions are filtered to those where all three annotators agree. The processed test sets contain 1,892 and 1,109 questions for the two parts respectively, with a slight class imbalance and 56.4\% and 61.7\% of the questions having the label `yes' respectively. 

\begin{table}[!htbp]
    \centering
    \small
    \begin{tabular}{l|c|c}
    \toprule
        Method & Unanimous & Majority votes \\
    \midrule
        \citet{lipping2022clotho} & 73.1 & 63.2 \\
        \midrule
        Uncalibrated & 64.0 & 58.8 \\
        Zero Input & 65.2 & 60.1 \\
        Gaussian Noise & 38.6 & 43.8 \\
        Prior-Matched & 61.1 & 58.5 \\
        \bottomrule
    \end{tabular}
    \caption{Experimental results on Clotho-AQA test set.}
    \label{tab:clotho-aqa}
\end{table}

\noindent We prompt Whisper in a similar fashion to the previous audio classification approach, however the input question is now used as the prompt for the decoder. As before, the audio clip is provided to the model's encoder, and the system likelihood of generating `yes' and `no' are used as class logits. The setup is depicted in Figure \ref{fig:aqa_method}. The baseline from \citet{lipping2022clotho} is a BiLSTM-based system with a binary classification head, trained in a supervised fashion on the labelled training corpus.

\begin{figure}[!htbp]
    \centering
    \includegraphics[width=\linewidth]{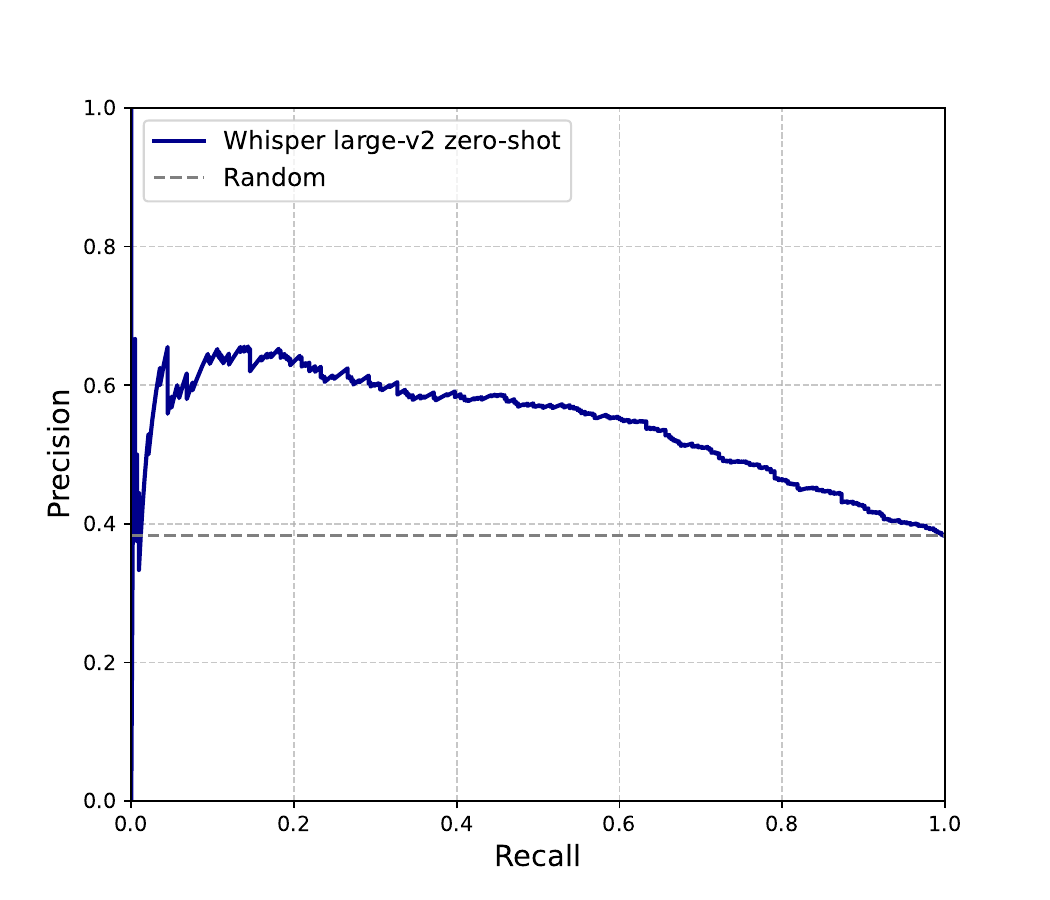}
    \caption{Precision-Recall curve for Whisper large-v2 prompted for Clotho-AQA. `no', the rarer event, is used as the positive class for detection.}
    \label{aqa_roc}
\end{figure}

\noindent Table \ref{tab:clotho-aqa} presents experimental results, where zero-shot Whisper achieves an accuracy of 64.0 for the unanimous test set. Note that due to class imbalance a system that always predicts `yes' will have an accuracy of 61.7\%. However, the precision of the proposed method is 65.9\% and 60.9\% for the `yes' and `no' decisions respectively, both significantly above random. Due to this inherent class imbalance, prior matching (which ensures the output prior is uniform) degrades performance and yields lower accuracy. Applying the null-input normalization techniques can improve performance with zero-input, although Gaussian noise harms performance (as it overcompensates the bias and makes predictions biased to predict mostly `no'). Similar observations are found when considering the `majority' processed test data.  

To confirm the extent to which Whisper is making informed, rather than random, decisions the precision and recall curve for the rarer class, `no' is shown in  Figure \ref{aqa_roc} on the unanimous set. It is clear that there is significant information in Whisper's zero-shot predictions and performance is notably better than random at all decision thresholds.



\section{Conclusions}
This paper is the first to examine the emergent ability of foundation ASR models on audio-classification tasks, that were not seen in training. Over a range of tasks, we show that zero-shot prompting of Whisper can yield effective performance. Calibration methods can be used to readjust the output distribution for better task alignment, allowing Whisper to achieve better performance compared to previous zero-shot works, and demonstrating its potential for cross-task generalization. 




\section*{Acknowledgements}
This work is supported by Cambridge University Press \& Assessment (CUP\&A), a department of The Chancellor, Masters, and Scholars of the University of Cambridge.  

\section{Limitations}
Prior-matching, which yielded considerable gains, assumes that the classes are fairly balanced and requires unlabelled in-domain data (or a large test set to be evaluated). This approach may not apply to settings where there are strong class imbalances, nor when little data is available.

\section{Ethical Considerations}
This is an introductory study that demonstrates that Whisper can be used for zero-shot audio classification tasks. However, the system may not generalize well to some tasks not considered in this paper. Our zero-shot method should be used with a level of caution, especially if leveraging the system for real-world applications.

\bibliography{anthology,custom}

\begin{thebibliography}{42}
\expandafter\ifx\csname natexlab\endcsname\relax\def\natexlab#1{#1}\fi

\bibitem[{Bang et~al.(2023)Bang, Cahyawijaya, Lee, Dai, Su, Wilie, Lovenia, Ji, Yu, Chung, Do, Xu, and Fung}]{bang2023multitask}
Yejin Bang, Samuel Cahyawijaya, Nayeon Lee, Wenliang Dai, Dan Su, Bryan Wilie, Holy Lovenia, Ziwei Ji, Tiezheng Yu, Willy Chung, Quyet~V. Do, Yan Xu, and Pascale Fung. 2023.
\newblock \href {https://arxiv.org/abs/2302.04023} {{A Multitask, Multilingual, Multimodal Evaluation of ChatGPT on Reasoning, Hallucination, and Interactivity}}.
\newblock \emph{arXiv preprint arXiv:2302.04023}.

\bibitem[{Bannò et~al.(2023)Bannò, Ma, Qian, Knill, and Gales}]{bannò2023endtoend}
Stefano Bannò, Rao Ma, Mengjie Qian, Kate~M. Knill, and Mark J.~F. Gales. 2023.
\newblock \href {https://doi.org/https://doi.org/10.48550/arXiv.2311.05550} {Towards end-to-end spoken grammatical error correction}.
\newblock \emph{arXiv preprint arXiv:2311.05550}.

\bibitem[{Barlow and Brunk(1972)}]{isotonic1972}
R.~E. Barlow and H.~D. Brunk. 1972.
\newblock \href {https://doi.org/10.1080/01621459.1972.10481216} {The isotonic regression problem and its dual}.
\newblock \emph{Journal of the American Statistical Association}, 67(337):140--147.

\bibitem[{Brown et~al.(2020)Brown, Mann, Ryder, Subbiah, Kaplan, Dhariwal, Neelakantan, Shyam, Sastry, Askell, Agarwal, Herbert-Voss, Krueger, Henighan, Child, Ramesh, Ziegler, Wu, Winter, Hesse, Chen, Sigler, Litwin, Gray, Chess, Clark, Berner, McCandlish, Radford, Sutskever, and Amodei}]{brown2020language}
Tom Brown, Benjamin Mann, Nick Ryder, Melanie Subbiah, Jared~D Kaplan, Prafulla Dhariwal, Arvind Neelakantan, Pranav Shyam, Girish Sastry, Amanda Askell, Sandhini Agarwal, Ariel Herbert-Voss, Gretchen Krueger, Tom Henighan, Rewon Child, Aditya Ramesh, Daniel Ziegler, Jeffrey Wu, Clemens Winter, Chris Hesse, Mark Chen, Eric Sigler, Mateusz Litwin, Scott Gray, Benjamin Chess, Jack Clark, Christopher Berner, Sam McCandlish, Alec Radford, Ilya Sutskever, and Dario Amodei. 2020.
\newblock \href {https://proceedings.neurips.cc/paper_files/paper/2020/file/1457c0d6bfcb4967418bfb8ac142f64a-Paper.pdf} {{Language Models are Few-Shot Learners}}.
\newblock In \emph{Advances in Neural Information Processing Systems}, volume~33, pages 1877--1901. Curran Associates, Inc.

\bibitem[{Cao et~al.(2014)Cao, Cooper, Keutmann, Gur, Nenkova, and Verma}]{cao2014crema}
Houwei Cao, David~G Cooper, Michael~K Keutmann, Ruben~C Gur, Ani Nenkova, and Ragini Verma. 2014.
\newblock {CREMA-D}: Crowd-sourced emotional multimodal actors dataset.
\newblock \emph{IEEE transactions on affective computing}, 5(4):377--390.

\bibitem[{Chung et~al.(2022)Chung, Hou, Longpre, Zoph, Tay, Fedus, Li, Wang, Dehghani, Brahma et~al.}]{chung2022scaling}
Hyung~Won Chung, Le~Hou, Shayne Longpre, Barret Zoph, Yi~Tay, William Fedus, Eric Li, Xuezhi Wang, Mostafa Dehghani, Siddhartha Brahma, et~al. 2022.
\newblock \href {https://doi.org/https://doi.org/10.48550/arXiv.2210.11416} {Scaling instruction-finetuned language models}.
\newblock \emph{arXiv preprint arXiv:2210.11416}.

\bibitem[{Elizalde et~al.(2023)Elizalde, Deshmukh, Al~Ismail, and Wang}]{elizalde2023clap}
Benjamin Elizalde, Soham Deshmukh, Mahmoud Al~Ismail, and Huaming Wang. 2023.
\newblock \href {https://doi.org/10.1109/ICASSP49357.2023.10095889} {{CLAP: Learning audio concepts from natural language supervision}}.
\newblock In \emph{ICASSP 2023-2023 IEEE International Conference on Acoustics, Speech and Signal Processing (ICASSP)}, pages 1--5. IEEE.

\bibitem[{Gao et~al.(2021)Gao, Fisch, and Chen}]{gao-etal-2021-making}
Tianyu Gao, Adam Fisch, and Danqi Chen. 2021.
\newblock \href {https://doi.org/10.18653/v1/2021.acl-long.295} {Making pre-trained language models better few-shot learners}.
\newblock In \emph{Proceedings of the 59th Annual Meeting of the Association for Computational Linguistics and the 11th International Joint Conference on Natural Language Processing (Volume 1: Long Papers)}, pages 3816--3830, Online. Association for Computational Linguistics.

\bibitem[{Gong et~al.(2023)Gong, Khurana, Karlinsky, and Glass}]{gong_whisperat}
Yuan Gong, Sameer Khurana, Leonid Karlinsky, and James Glass. 2023.
\newblock \href {https://doi.org/10.21437/Interspeech.2023-2193} {{Whisper-AT: Noise-Robust Automatic Speech Recognizers are Also Strong General Audio Event Taggers}}.
\newblock In \emph{Proc. INTERSPEECH 2023}, pages 2798--2802.

\bibitem[{Gong et~al.(2022)Gong, Yu, and Glass}]{gong2022vocalsound}
Yuan Gong, Jin Yu, and James Glass. 2022.
\newblock Vocalsound: A dataset for improving human vocal sounds recognition.
\newblock In \emph{ICASSP 2022-2022 IEEE International Conference on Acoustics, Speech and Signal Processing (ICASSP)}, pages 151--155. IEEE.

\bibitem[{Graves et~al.(2006)Graves, Fern{\'a}ndez, Gomez, and Schmidhuber}]{graves2006connectionist}
Alex Graves, Santiago Fern{\'a}ndez, Faustino Gomez, and J{\"u}rgen Schmidhuber. 2006.
\newblock {Connectionist temporal classification: labelling unsegmented sequence data with recurrent neural networks}.
\newblock In \emph{Proceedings of the 23rd international conference on Machine learning}, pages 369--376.

\bibitem[{Guo et~al.(2017)Guo, Pleiss, Sun, and Weinberger}]{guo2017calibration}
Chuan Guo, Geoff Pleiss, Yu~Sun, and Kilian~Q Weinberger. 2017.
\newblock \href {https://proceedings.mlr.press/v70/guo17a/guo17a.pdf} {On calibration of modern neural networks}.
\newblock In \emph{Proc. of the 34th International Conference on Machine Learning (ICML)}, pages 1321--1330. PMLR.

\bibitem[{Guo et~al.(2022)Guo, Yang, and Abbasi}]{guo-etal-2022-auto}
Yue Guo, Yi~Yang, and Ahmed Abbasi. 2022.
\newblock \href {https://doi.org/10.18653/v1/2022.acl-long.72} {Auto-debias: Debiasing masked language models with automated biased prompts}.
\newblock In \emph{Proceedings of the 60th Annual Meeting of the Association for Computational Linguistics (Volume 1: Long Papers)}, pages 1012--1023, Dublin, Ireland. Association for Computational Linguistics.

\bibitem[{Guzhov et~al.(2022)Guzhov, Raue, Hees, and Dengel}]{guzhov2022audioclip}
Andrey Guzhov, Federico Raue, J{\"o}rn Hees, and Andreas Dengel. 2022.
\newblock \href {https://doi.org/10.1109/ICASSP43922.2022.9747631} {{AudioCLIP}: Extending clip to image, text and audio}.
\newblock In \emph{ICASSP 2022-2022 IEEE International Conference on Acoustics, Speech and Signal Processing (ICASSP)}, pages 976--980. IEEE.

\bibitem[{Hu et~al.(2021)Hu, Wallis, Allen-Zhu, Li, Wang, Wang, Chen et~al.}]{hu2021lora}
Edward~J Hu, Phillip Wallis, Zeyuan Allen-Zhu, Yuanzhi Li, Shean Wang, Lu~Wang, Weizhu Chen, et~al. 2021.
\newblock {LoRA}: Low-rank adaptation of large language models.
\newblock In \emph{International Conference on Learning Representations}.

\bibitem[{Kojima et~al.(2022)Kojima, Gu, Reid, Matsuo, and Iwasawa}]{kojima2022large}
Takeshi Kojima, Shixiang~Shane Gu, Machel Reid, Yutaka Matsuo, and Yusuke Iwasawa. 2022.
\newblock Large language models are zero-shot reasoners.
\newblock \emph{Advances in neural information processing systems}, 35:22199--22213.

\bibitem[{Lester et~al.(2021)Lester, Al-Rfou, and Constant}]{lester2021power}
Brian Lester, Rami Al-Rfou, and Noah Constant. 2021.
\newblock The power of scale for parameter-efficient prompt tuning.
\newblock In \emph{Proceedings of the 2021 Conference on Empirical Methods in Natural Language Processing}, pages 3045--3059.

\bibitem[{Li et~al.(2017)Li, Jabri, Joulin, and van~der Maaten}]{li2017learning}
Ang Li, Allan Jabri, Armand Joulin, and Laurens van~der Maaten. 2017.
\newblock \href {https://doi.org/10.1109/ICCV.2017.449} {{Learning Visual N-Grams from Web Data}}.
\newblock In \emph{IEEE International Conference on Computer Vision (ICCV)}, pages 4193--4202.

\bibitem[{Lipping et~al.(2022)Lipping, Sudarsanam, Drossos, and Virtanen}]{lipping2022clotho}
Samuel Lipping, Parthasaarathy Sudarsanam, Konstantinos Drossos, and Tuomas Virtanen. 2022.
\newblock \href {https://eurasip.org/Proceedings/Eusipco/Eusipco2022/pdfs/0001140.pdf} {{Clotho-AQA}: A crowdsourced dataset for audio question answering}.
\newblock In \emph{2022 30th European Signal Processing Conference (EUSIPCO)}, pages 1140--1144.

\bibitem[{Liusie et~al.(2023)Liusie, Manakul, and Gales}]{liusie2023mitigating}
Adian Liusie, Potsawee Manakul, and Mark~JF Gales. 2023.
\newblock \href {http://www.afnlp.org/conferences/ijcnlp2023/proceedings/main-findings/cdrom/pdf/2023.findings-ijcnlp.29.pdf} {{Mitigating Word Bias in Zero-shot Prompt-based Classifiers}}.
\newblock In \emph{Proc. of the 13th International Joint Conference on Natural Language Processing and the 3rd Conference of the Asia-Pacific Chapter of the Association for Computational Linguistics (Findings Papers)}, pages 327–--335.

\bibitem[{Luna-Jim{\'e}nez et~al.(2021)Luna-Jim{\'e}nez, Kleinlein, Griol, Callejas, Montero, and Fern{\'a}ndez-Mart{\'\i}nez}]{luna2021proposal}
Cristina Luna-Jim{\'e}nez, Ricardo Kleinlein, David Griol, Zoraida Callejas, Juan~M Montero, and Fernando Fern{\'a}ndez-Mart{\'\i}nez. 2021.
\newblock A proposal for multimodal emotion recognition using aural transformers and action units on {RAVDESS} dataset.
\newblock \emph{Applied Sciences}, 12(1):327.

\bibitem[{Ma et~al.(2023)Ma, Qian, Gales, and Knill}]{ma2023adapting}
Rao Ma, Mengjie Qian, Mark~JF Gales, and Kate~M Knill. 2023.
\newblock Adapting an asr foundation model for spoken language assessment.
\newblock \emph{arXiv preprint arXiv:2307.09378}.

\bibitem[{Mesaros et~al.(2016)Mesaros, Heittola, and Virtanen}]{mesaros2016tut}
Annamaria Mesaros, Toni Heittola, and Tuomas Virtanen. 2016.
\newblock {TUT} database for acoustic scene classification and sound event detection.
\newblock In \emph{2016 24th European Signal Processing Conference (EUSIPCO)}, pages 1128--1132. IEEE.

\bibitem[{Ouyang et~al.(2022)Ouyang, Wu, Jiang, Almeida, Wainwright, Mishkin, Zhang, Agarwal, Slama, Ray et~al.}]{ouyang2022training}
Long Ouyang, Jeffrey Wu, Xu~Jiang, Diogo Almeida, Carroll Wainwright, Pamela Mishkin, Chong Zhang, Sandhini Agarwal, Katarina Slama, Alex Ray, et~al. 2022.
\newblock \href {https://proceedings.neurips.cc/paper_files/paper/2022/file/b1efde53be364a73914f58805a001731-Paper-Conference.pdf} {Training language models to follow instructions with human feedback}.
\newblock \emph{Advances in Neural Information Processing Systems (NeurIPS 2022)}, 35:27730--27744.

\bibitem[{Peng et~al.(2023)Peng, Yan, Watanabe, and Harwath}]{peng2023prompting}
Puyuan Peng, Brian Yan, Shinji Watanabe, and David Harwath. 2023.
\newblock \href {https://doi.org/10.48550/arXiv.2305.11095} {Prompting the hidden talent of web-scale speech models for zero-shot task generalization}.
\newblock \emph{arXiv preprint arXiv:2305.11095}.

\bibitem[{Piczak(2015)}]{piczak2015esc}
Karol~J Piczak. 2015.
\newblock Esc: Dataset for environmental sound classification.
\newblock In \emph{Proceedings of the 23rd ACM international conference on Multimedia}, pages 1015--1018.

\bibitem[{Platt(1999)}]{platt1999probabilistic}
John Platt. 1999.
\newblock Probabilistic outputs for support vector machines and comparisons to regularized likelihood methods.
\newblock In \emph{Advances in Large Margin Classifiers}. MIT Press, Cambridge MA.

\bibitem[{Pratap et~al.(2023)Pratap, Tjandra, Shi, Tomasello, Babu, Kundu, Elkahky, Ni, Vyas, Fazel-Zarandi, Baevski, Adi, Zhang, Hsu, Conneau, and Auli}]{pratap2023scaling}
Vineel Pratap, Andros Tjandra, Bowen Shi, Paden Tomasello, Arun Babu, Sayani Kundu, Ali Elkahky, Zhaoheng Ni, Apoorv Vyas, Maryam Fazel-Zarandi, Alexei Baevski, Yossi Adi, Xiaohui Zhang, Wei-Ning Hsu, Alexis Conneau, and Michael Auli. 2023.
\newblock \href {https://doi.org/10.48550/arXiv.2305.13516} {Scaling speech technology to 1,000+ languages}.
\newblock \emph{arXiv preprint arXiv:2305.13516}.

\bibitem[{Radford et~al.(2021)Radford, Kim, Hallacy, Ramesh, Goh, Agarwal, Sastry, Askell, Mishkin, Clark et~al.}]{radford2021learning}
Alec Radford, Jong~Wook Kim, Chris Hallacy, Aditya Ramesh, Gabriel Goh, Sandhini Agarwal, Girish Sastry, Amanda Askell, Pamela Mishkin, Jack Clark, et~al. 2021.
\newblock \href {http://proceedings.mlr.press/v139/radford21a/radford21a.pdf} {Learning transferable visual models from natural language supervision}.
\newblock In \emph{Proc. of the 38th International Conference on Machine Learning}, pages 8748--8763. PMLR.

\bibitem[{Radford et~al.(2023)Radford, Kim, Xu, Brockman, McLeavey, and Sutskever}]{radford2023robust}
Alec Radford, Jong~Wook Kim, Tao Xu, Greg Brockman, Christine McLeavey, and Ilya Sutskever. 2023.
\newblock \href {https://proceedings.mlr.press/v202/radford23a/radford23a.pdf} {Robust speech recognition via large-scale weak supervision}.
\newblock In \emph{Proc. of the 40th International Conference on Machine Learning}, pages 28492--28518. PMLR.

\bibitem[{Radford et~al.(2019)Radford, Wu, Child, Luan, Amodei, Sutskever et~al.}]{radford2019language}
Alec Radford, Jeffrey Wu, Rewon Child, David Luan, Dario Amodei, Ilya Sutskever, et~al. 2019.
\newblock Language models are unsupervised multitask learners.
\newblock \emph{OpenAI blog}, 1(8):9.

\bibitem[{Salamon et~al.(2014)Salamon, Jacoby, and Bello}]{salamon2014dataset}
Justin Salamon, Christopher Jacoby, and Juan~Pablo Bello. 2014.
\newblock A dataset and taxonomy for urban sound research.
\newblock In \emph{Proceedings of the 22nd ACM international conference on Multimedia}, pages 1041--1044.

\bibitem[{Sanh et~al.(2022)Sanh, Webson, Raffel, Bach, Sutawika, Alyafeai, Chaffin, Stiegler, Raja, Dey et~al.}]{sanh2021multitask}
Victor Sanh, Albert Webson, Colin Raffel, Stephen Bach, Lintang Sutawika, Zaid Alyafeai, Antoine Chaffin, Arnaud Stiegler, Arun Raja, Manan Dey, et~al. 2022.
\newblock \href {https://openreview.net/pdf?id=9Vrb9D0WI4} {Multitask prompted training enables zero-shot task generalization}.
\newblock In \emph{International Conference on Learning Representations (ICLR) 2022}.

\bibitem[{Schaeffer et~al.(2023)Schaeffer, Miranda, and Koyejo}]{schaeffer2023emergent}
Rylan Schaeffer, Brando Miranda, and Sanmi Koyejo. 2023.
\newblock \href {https://openreview.net/pdf?id=JRdN9GcI52} {Are emergent abilities of large language models a mirage?}
\newblock In \emph{Challenges in Deployable Generative AI Workshop at ICML}.

\bibitem[{Schick and Sch{\"u}tze(2021)}]{schick-schutze-2021-exploiting}
Timo Schick and Hinrich Sch{\"u}tze. 2021.
\newblock \href {https://doi.org/10.18653/v1/2021.eacl-main.20} {Exploiting cloze-questions for few-shot text classification and natural language inference}.
\newblock In \emph{Proceedings of the 16th Conference of the European Chapter of the Association for Computational Linguistics: Main Volume}, pages 255--269, Online. Association for Computational Linguistics.

\bibitem[{St{\"o}ter et~al.(2018)St{\"o}ter, Chakrabarty, Habets, and Edler}]{stoter2018libricount}
Fabian-Robert St{\"o}ter, Soumitro Chakrabarty, Emanu{\"e}l Habets, and Bernd Edler. 2018.
\newblock {LibriCount}, a dataset for speaker count estimation.

\bibitem[{Sturm(2013)}]{sturm2013gtzan}
Bob~L Sturm. 2013.
\newblock The {GTZAN} dataset: Its contents, its faults, their effects on evaluation, and its future use.
\newblock \emph{arXiv preprint arXiv:1306.1461}.

\bibitem[{Touvron et~al.(2023)Touvron, Lavril, Izacard, Martinet, Lachaux, Lacroix, Rozi{\`e}re, Goyal, Hambro, Azhar et~al.}]{touvron2023llama}
Hugo Touvron, Thibaut Lavril, Gautier Izacard, Xavier Martinet, Marie-Anne Lachaux, Timoth{\'e}e Lacroix, Baptiste Rozi{\`e}re, Naman Goyal, Eric Hambro, Faisal Azhar, et~al. 2023.
\newblock \href {https://arxiv.org/pdf/2302.13971.pdf} {{LLaMA}: Open and efficient foundation language models}.
\newblock \emph{arXiv preprint arXiv:2302.13971}.

\bibitem[{Wang et~al.(2023{\natexlab{a}})Wang, Li, Guo, Qiao, Li, Shang, Wei, Tao, Zhang, and Yang}]{wang2023whislu}
Minghan Wang, Yinglu Li, Jiaxin Guo, Xiaosong Qiao, Zongyao Li, Hengchao Shang, Daimeng Wei, Shimin Tao, Min Zhang, and Hao Yang. 2023{\natexlab{a}}.
\newblock \href {https://doi.org/10.21437/Interspeech.2023-1505} {{WhiSLU: End-to-End Spoken Language Understanding with Whisper}}.
\newblock In \emph{Proc. INTERSPEECH 2023}, pages 770--774.

\bibitem[{Wang et~al.(2023{\natexlab{b}})Wang, Yang, Wu, and Zhang}]{wang2023can}
Siyin Wang, Chao-Han~Huck Yang, Ji~Wu, and Chao Zhang. 2023{\natexlab{b}}.
\newblock Can whisper perform speech-based in-context learning.
\newblock \emph{arXiv preprint arXiv:2309.07081}.

\bibitem[{Wei et~al.(2022)Wei, Tay, Bommasani, Raffel, Zoph, Borgeaud, Yogatama, Bosma, Zhou, Metzler, Chi, Hashimoto, Vinyals, Liang, Dean, and Fedus}]{wei2022emergent}
Jason Wei, Yi~Tay, Rishi Bommasani, Colin Raffel, Barret Zoph, Sebastian Borgeaud, Dani Yogatama, Maarten Bosma, Denny Zhou, Donald Metzler, Ed~H. Chi, Tatsunori Hashimoto, Oriol Vinyals, Percy Liang, Jeff Dean, and William Fedus. 2022.
\newblock \href {https://openreview.net/forum?id=yzkSU5zdwD} {Emergent abilities of large language models}.
\newblock \emph{Transactions on Machine Learning Research}.
\newblock Survey Certification.

\bibitem[{Zhao et~al.(2021)Zhao, Wallace, Feng, Klein, and Singh}]{zhao2021calibrate}
Zihao Zhao, Eric Wallace, Shi Feng, Dan Klein, and Sameer Singh. 2021.
\newblock \href {http://proceedings.mlr.press/v139/zhao21c/zhao21c.pdf} {Calibrate before use: Improving few-shot performance of language models}.
\newblock In \emph{Proc. of the 38th International Conference on Machine Learning}, pages 12697--12706. PMLR.

\end{thebibliography}
\newpage
\onecolumn
\appendix

\section{Full Results}
\label{sec:appendix_main}

\begin{table}[!h]
  \centering \footnotesize
   \renewcommand\tabcolsep{5pt}
\begin{tabular}{l|c|c|c|c|c|c|c|c|c}
        \toprule
        Model & ESC50 & US8K  & TUT & Vocal & RAVDESS & CREMA-D & GTZAN & LibriCount & \textbf{Avg.} \\
        \midrule
        \rowcolor{Gray}
        \multicolumn{5}{c}{Baselines (\S \ref{sec:baselines})} & \multicolumn{5}{l}{}  \\        
        Random & 2.0 & 10.0 & 6.7 & 16.7 & 12.5 & 16.7 & 10.0 & 9.1 & 10.4 \\
        AudioCLIP & 69.4 & 65.3 & - & - & - & - & - & - & - \\
        CLAP  & 82.6 & 73.2 & 29.6 & 49.4 & 16.0 & 17.8 & 25.2 & 17.9 & 39.0 \\
        \rowcolor{Gray}
        \multicolumn{5}{c}{Uncalibrated (\S \ref{sec:default})} & \multicolumn{5}{l}{}  \\        
        MMS large (1B) & 1.7  & 9.6  & 4.9 & 14.2 & 13.5 & 17.2 & 8.3  & 8.4 & 9.7  \\
        Whisper tiny.en (39M) & 3.7 & 16.4 & 6.7 & 16.7 & 13.3 & 17.4 & 13.9 & 9.3 & 12.2 \\
        Whisper tiny (39M) & 4.2 & 12.9 & 6.5 & 17.0 & 12.4 & 15.9 & 13.3 & 7.8 & 11.3 \\
        Whisper base.en (74M) & 5.9 & 20.4 & 6.6 & 35.1 & 13.2 & 16.0 & 13.6 & 10.2 & 15.1 \\
        Whisper base (74M) & 6.8 & 23.7 & 6.6 & 39.0 & 14.9 & 16.3 & 21.7 & 9.5 & 17.3 \\
        Whisper small.en (244M) & 10.3 & 41.9 & 7.0 & 45.0 & 14.7 & 14.8 & 14.6 & 7.2 & 19.4 \\
        Whisper small (244M) & 21.0 & 39.3 & 8.2 & 46.6 & 15.5 & 18.9 & 23.7 & 9.2 & 22.8 \\
        Whisper medium.en (769M) & 27.9 & 39.5 & 7.2 & 59.0 & 15.3 & 20.9 & 15.2 & 8.2 & 24.2 \\ 
        Whisper medium (769M)    & 29.7 & 45.8 & 7.5 & 44.6 & 16.7 & 19.9 & 28.4 & 9.4 & 25.2 \\ 
        Whisper large-v1 (1.6B) & 33.7 & 44.8 & 8.3 & 58.2 & 15.0 & 21.6 & 35.2 & 8.2 & 28.2 \\
        Whisper large-v2 (1.6B)  & 38.9 & 50.5 & 7.7 & 60.1 & 15.1 & 20.2 & 38.2 & 9.2 & 30.0 \\
        Whisper large-v3 (1.6B)  & 12.0 & 38.3 & 7.0 & 43.0 & 13.6 & 19.5 & 14.4 & 9.3 & 19.6 \\
        \rowcolor{Gray}
        \multicolumn{5}{c}{Zero-Input (\S \ref{sec:null_input})} & \multicolumn{5}{l}{}  \\        
        MMS large (1B)  & 2.2 & 11.7 & 4.2 & 16.5 & 12.1 & 15.9 & 7.5 & 10.0 & 10.0 \\
        Whisper tiny.en (39M) & 12.7 & 19.7 & 7.5 & 30.9 & 20.6 & 18.9 & 12.8 & 9.4 & 16.6 \\
        Whisper tiny (39M) & 10.5 & 24.2 & 7.7 & 28.0 & 15.8 & 17.7 & 17.7 & 7.9 & 16.2 \\
        Whisper base.en (74M) & 18.9 & 37.6 & 14.2 & 50.9 & 18.8 & 21.5 & 13.6 & 8.8 & 23.0 \\
        Whisper base (74M) & 19.4 & 36.2 & 12.1 & 52.7 & 14.4 & 16.5 & 17.5 & 11.1 & 22.5 \\
        Whisper small.en (244M) & 30.5 & 47.3 & 11.4 & 65.8 & 14.4 & 18.5 & 9.4 & 6.2 & 25.4 \\
        Whisper small (244M) & 30.9 & 41.0 & 19.8 & 54.3 & 14.7 & 17.2 & 38.8 & 10.1 & 28.4 \\
        Whisper medium.en (769M) & 44.1 & 53.3 & 21.5 & 57.2 & 20.1 & 21.2 & 12.2 & 8.6 & 29.8 \\ 
        Whisper medium (769M)    & 45.6 & 57.1 & 19.6 & 67.8 & 23.3 & 22.1 & 24.1 & 18.5 & 34.8 \\ 
        Whisper large-v1 (1.6B) & 47.1 & 58.5 & 24.9 & 59.3 & 18.5 & 26.0 & 32.8 & 8.7 & 34.5 \\
        Whisper large-v2 (1.6B)  & 35.9 & 52.1 & 18.0 & 57.5 & 29.4 & 26.5 & 45.8 & 13.6 & 34.9 \\
        Whisper large-v3 (1.6B)  & 23.9 & 38.4 & 21.2 & 60.9 & 15.7 & 20.7 & 11.8 & 13.9 & 25.8 \\
        \rowcolor{Gray}
        \multicolumn{5}{c}{Gaussian-Noise (\S \ref{sec:null_input})} & \multicolumn{5}{l}{}  \\        
        MMS large (1B)  & 2.4 & 12.6 & 7.9 & 13.0 & 12.7 & 17.0 & 14.9 & 11.9 & 11.5 \\
        Whisper tiny.en (39M) & 8.9 & 20.9 & 9.6 & 18.9 & 17.6 & 20.4 & 14.2 & 8.4 & 14.9 \\
        Whisper tiny (39M) & 5.8 & 19.4 & 11.8 & 16.7 & 13.5 & 17.1 & 16.4 & 7.7 & 13.6 \\
        Whisper base.en (74M) & 13.6 & 29.0 & 7.7 & 25.2 & 15.3 & 19.6 & 11.7 & 10.2 & 16.5 \\
        Whisper base (74M) & 17.5 & 27.6 & 6.5 & 39.5 & 12.8 & 17.8 & 12.2 & 9.0 & 17.9 \\
        Whisper small.en (244M) & 29.8 & 42.0 & 13.6 & 59.5 & 13.1 & 17.1 & 11.6 & 8.9 & 24.5 \\
        Whisper small (244M) & 31.2 & 49.0 & 14.8 & 52.5 & 24.0 & 21.4 & 41.6 & 12.6 & 30.9 \\
        Whisper medium.en (769M) & 36.8 & 45.8 & 20.0 & 68.9 & 17.2 & 20.4 & 10.0 & 8.9 & 28.5 \\ 
        Whisper medium (769M)    & 38.3 & 47.1 & 15.9 & 63.0 & 16.2 & 20.4 & 18.6 & 16.4 & 29.5 \\ 
        Whisper large-v1 (1.6B) & 47.9 & 58.7 & 26.1 & 44.8 & 18.7 & 20.1 & 20.5 & 9.1 & 30.7 \\
        Whisper large-v2 (1.6B)  & 43.8 & 53.2 & 22.1 & 62.4 & 20.4 & 18.8 & 50.8 & 15.0 & 35.8 \\
        Whisper large-v3 (1.6B)  & 22.9 & 29.3 & 14.1 & 43.1 & 16.5 & 17.6 & 19.4 & 14.9 & 22.2 \\
        \rowcolor{Gray}
        \multicolumn{5}{c}{Prior-matched (\S \ref{sec:prior_match})} & \multicolumn{5}{l}{}  \\        
        MMS large (1B) & 2.4  & 10.9 & 7.6 & 11.5 & 12.2 & 17.2 & 10.5 & 11.5 & 10.5  \\
        Whisper tiny.en (39M) & 17.3 & 30.4 & 11.7 & 41.5 & 19.6 & 20.4 & 19.3 & 8.8 & 21.1 \\
        Whisper tiny (39M) & 14.1 & 28.5 & 11.1 & 36.7 & 17.6 & 17.1 & 25.0 & 8.0 & 19.8 \\
        Whisper base.en (74M) & 24.6 & 46.2 & 11.7 & 58.6 & 20.3 & 20.1 & 25.4 & 12.3 & 27.4 \\
        Whisper base (74M) & 25.7 & 35.8 & 11.0 & 58.0 & 18.1 & 17.5 & 22.9 & 10.3 & 24.9 \\
        Whisper small.en (244M) & 43.7 & 55.5 & 15.7 & 78.8 & 24.6 & 18.7 & 28.1 & 7.7 & 34.1 \\
        Whisper small (244M) & 40.7 & 57.1 & 20.0 & 62.7 & 32.2 & 23.8 & 48.3 & 12.7 & 37.2 \\
        Whisper medium.en (769M) & 56.2 & 60.9 & 18.3 & 82.8 & 29.0 & 22.6 & 29.7 & 9.8  & 38.7 \\ 
        Whisper medium (769M)    & 57.5 & 61.6 & 25.2 & 82.4 & 35.0 & 25.9 & 48.6 & 16.3 & 44.1 \\ 
        Whisper large-v1 (1.6B) & 62.9 & 65.7 & 28.3 & 85.6 & 35.1 & 24.4 & 54.7 & 7.3 & 45.5 \\
        Whisper large-v2 (1.6B)  & 65.4 & 60.4 & 26.0 & 84.9 & 41.7 & 28.8 & 60.9 & 17.3 & \textbf{48.2} \\
        Whisper large-v3 (1.6B)  & 33.8 & 43.3 & 22.3 & 69.1 & 31.3 & 23.7 & 33.7 & 17.0 & 34.3 \\
        \bottomrule
    \end{tabular}
    \vspace{-4pt}
  \caption{Baseline and zero-shot task performance using the default prompt.}
  \label{fig:more_rst}
\end{table}

\noindent Table \ref{fig:more_rst} extends Table \ref{tab:main_table} and displays the zero-shot audio classification performance of different versions of the released ASR foundation models. As the results show, Whisper always exhibits better performance than random predictions, indicating that the model acquires the general ability of audio understanding when pre-trained on large-scale datasets. Null-input and prior matching calibration methods consistently improve the classification accuracy on selected tasks. All three Whisper large models share the same structure while the training strategy is slightly different. Compared to large-v1, Whisper large-v2 is trained on the data for 2.5 times more epochs with regularization techniques, leading to better audio classification accuracy. Nevertheless, the newly released Whisper large-v3 model shows inferior performance, which is trained on the combination of 1 million hours of weakly-labelled audio and 4 million hours of audio with pseudo labels decoded by large-v2. Results suggest that including pseudo-speech data harms the model's emergent ability for audio classification.

\section{Accuracy against Parameter Size}
\label{sec:appendix_acc}
\begin{figure*}[!h]
    \centering
    \includegraphics[width=0.24\columnwidth]{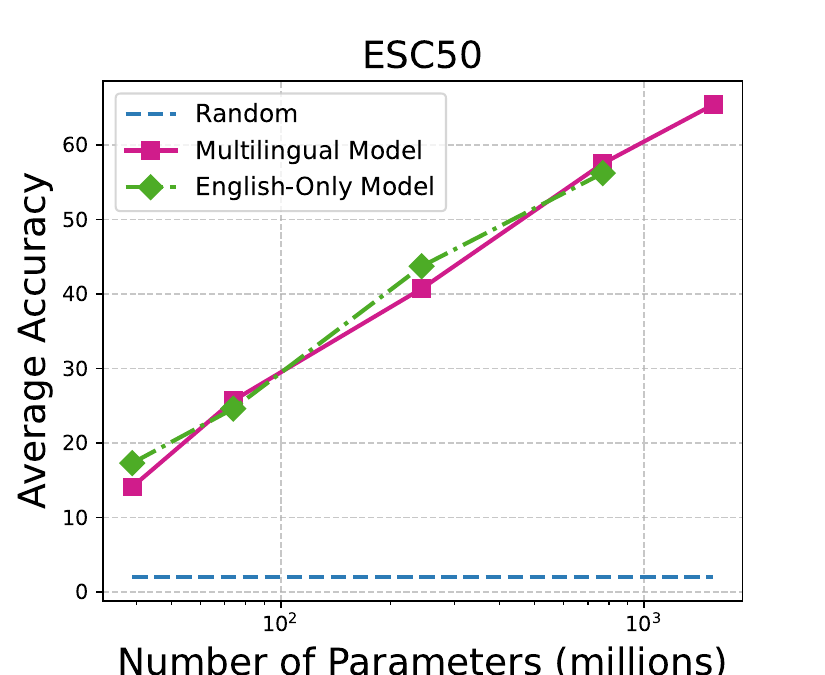}
    \includegraphics[width=0.24\columnwidth]{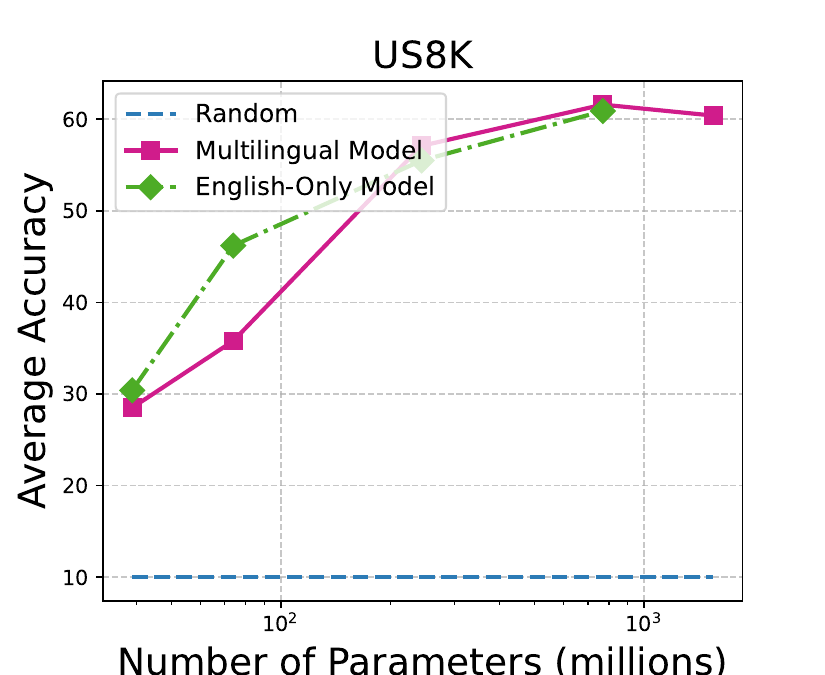}
    \includegraphics[width=0.24\columnwidth]{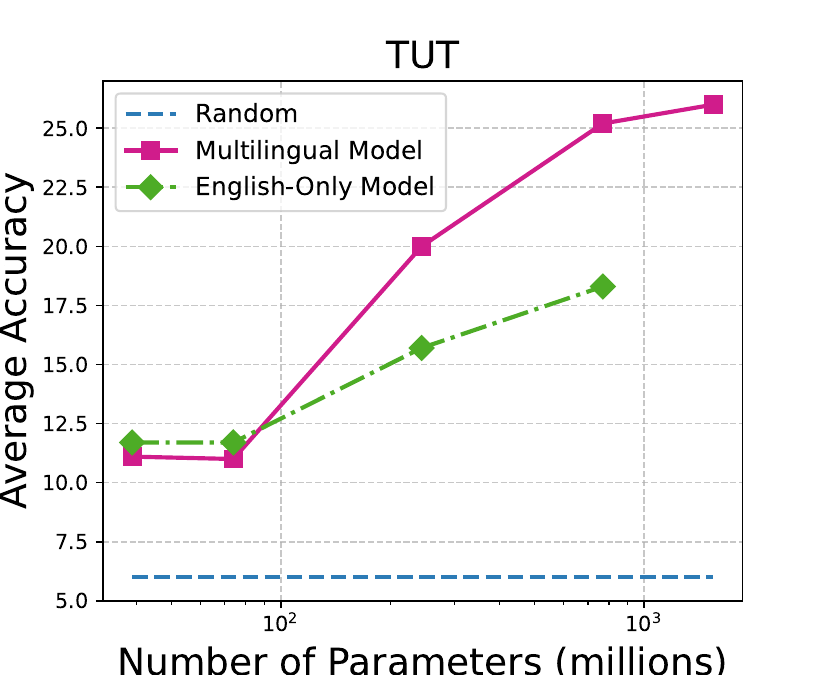}
    \includegraphics[width=0.24\columnwidth]{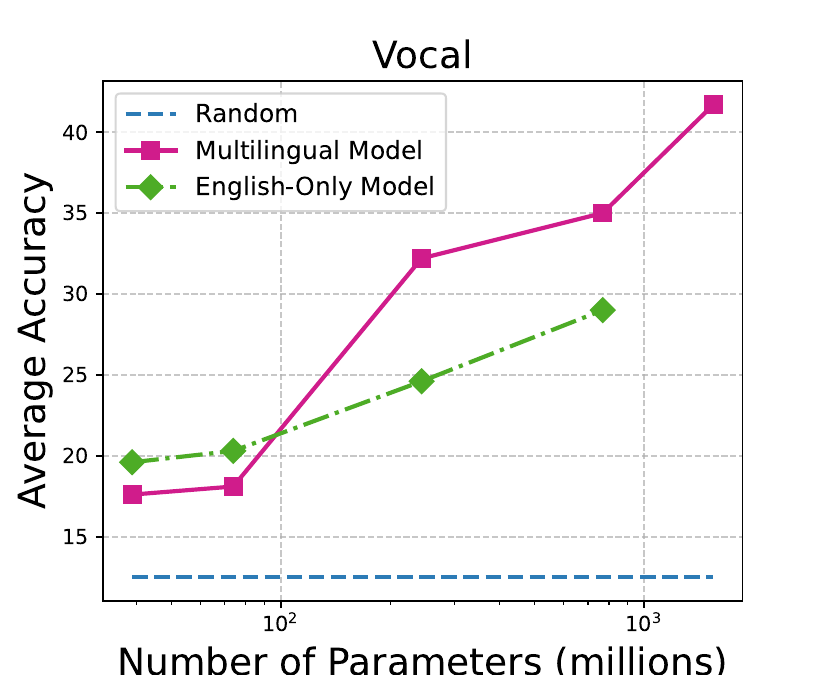}
    \includegraphics[width=0.24\columnwidth]{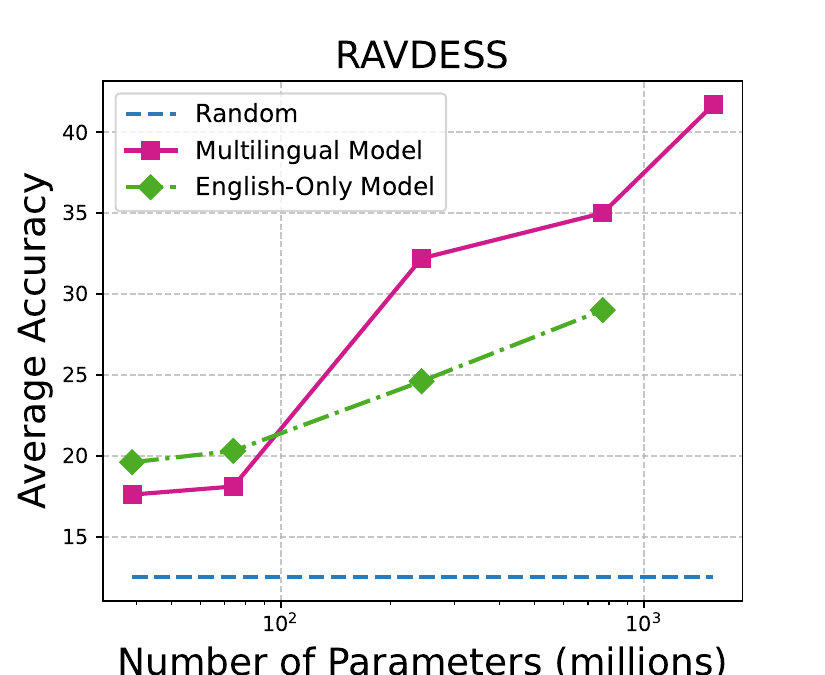}
    \includegraphics[width=0.24\columnwidth]{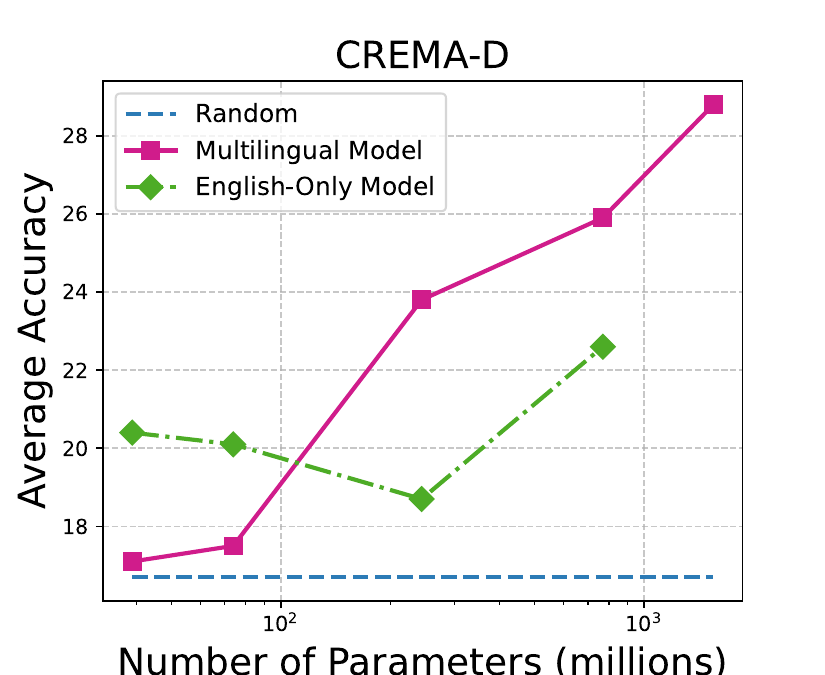}
    \includegraphics[width=0.24\columnwidth]{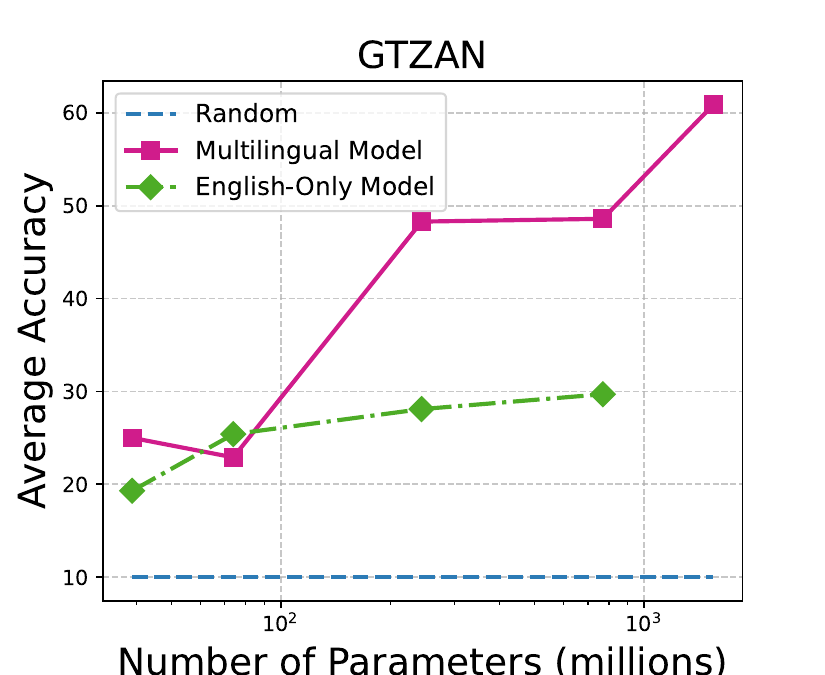}
    \includegraphics[width=0.24\columnwidth]{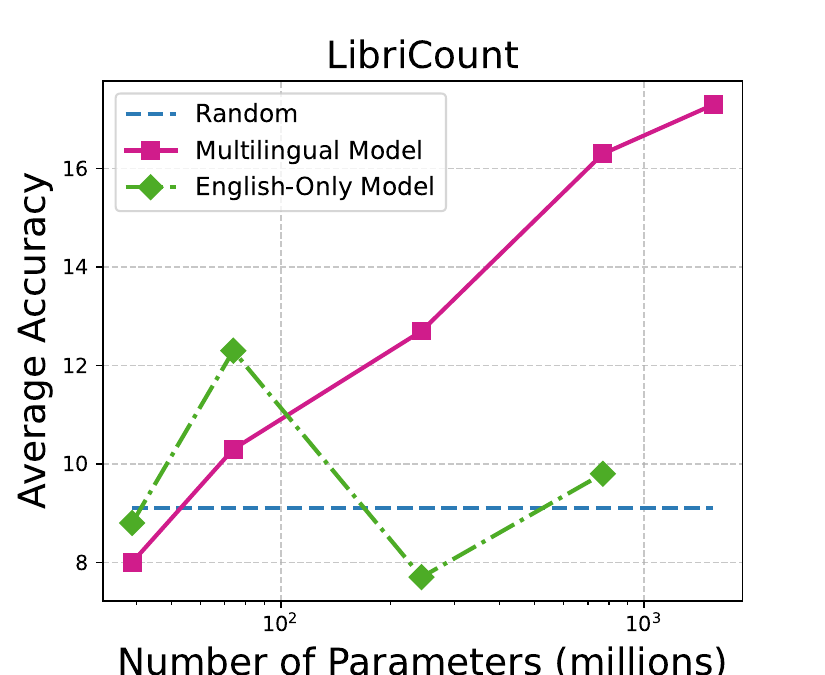}
    \caption{Accuracy on individual audio classification tasks across different sizes of Whisper models.}
    \label{fig:all_emergent}
\end{figure*}

\noindent Figure \ref{fig:all_emergent} shows the performance improvement of Whisper for various sizes, for both the English-only and the multilingual systems. In general, we observe better performance as the model size increases. For many tasks, we observe that as the number of parameters increases, the multilingual systems begin to outperform the English-only systems. However, for some tasks such as ESC50 and US8K, we observe comparable performance for the two systems over all model sizes. 

\section{Distribution of Predicted Classes}
\label{sec:appendix_pred}
\begin{figure*}[!h]
    \centering
    \includegraphics[width=0.24\columnwidth]{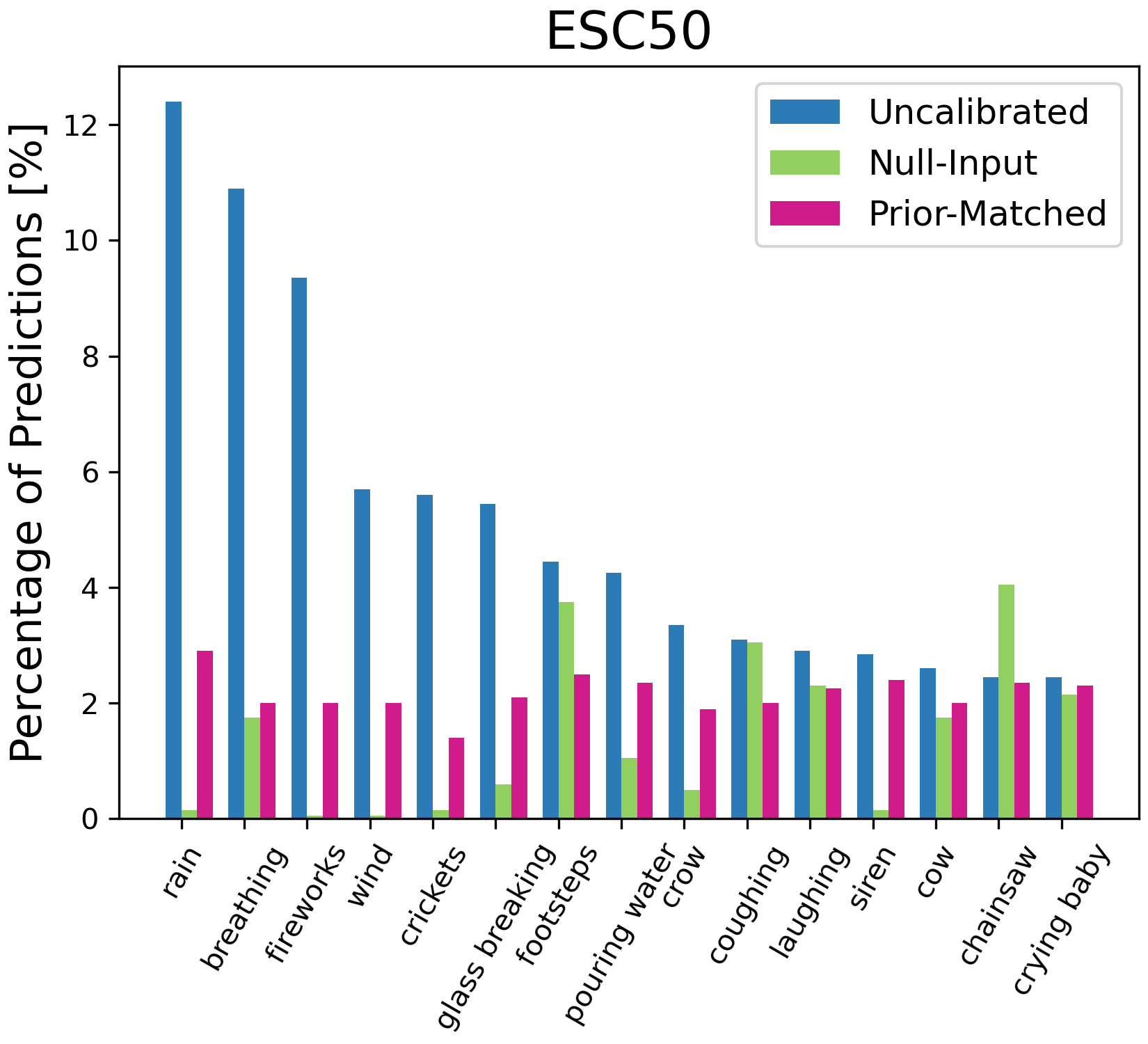}
    \includegraphics[width=0.24\columnwidth]{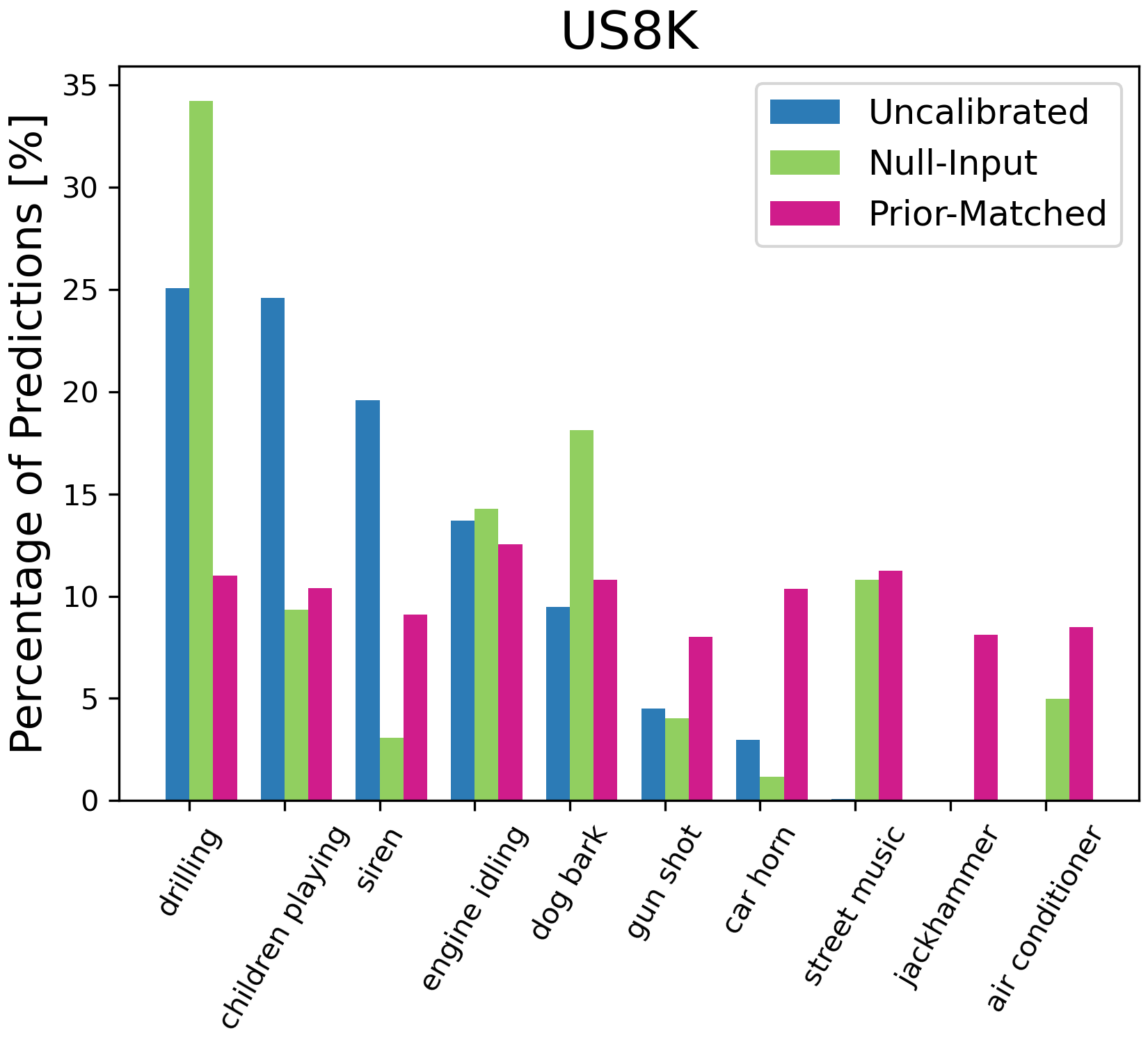}
    \includegraphics[width=0.24\columnwidth]{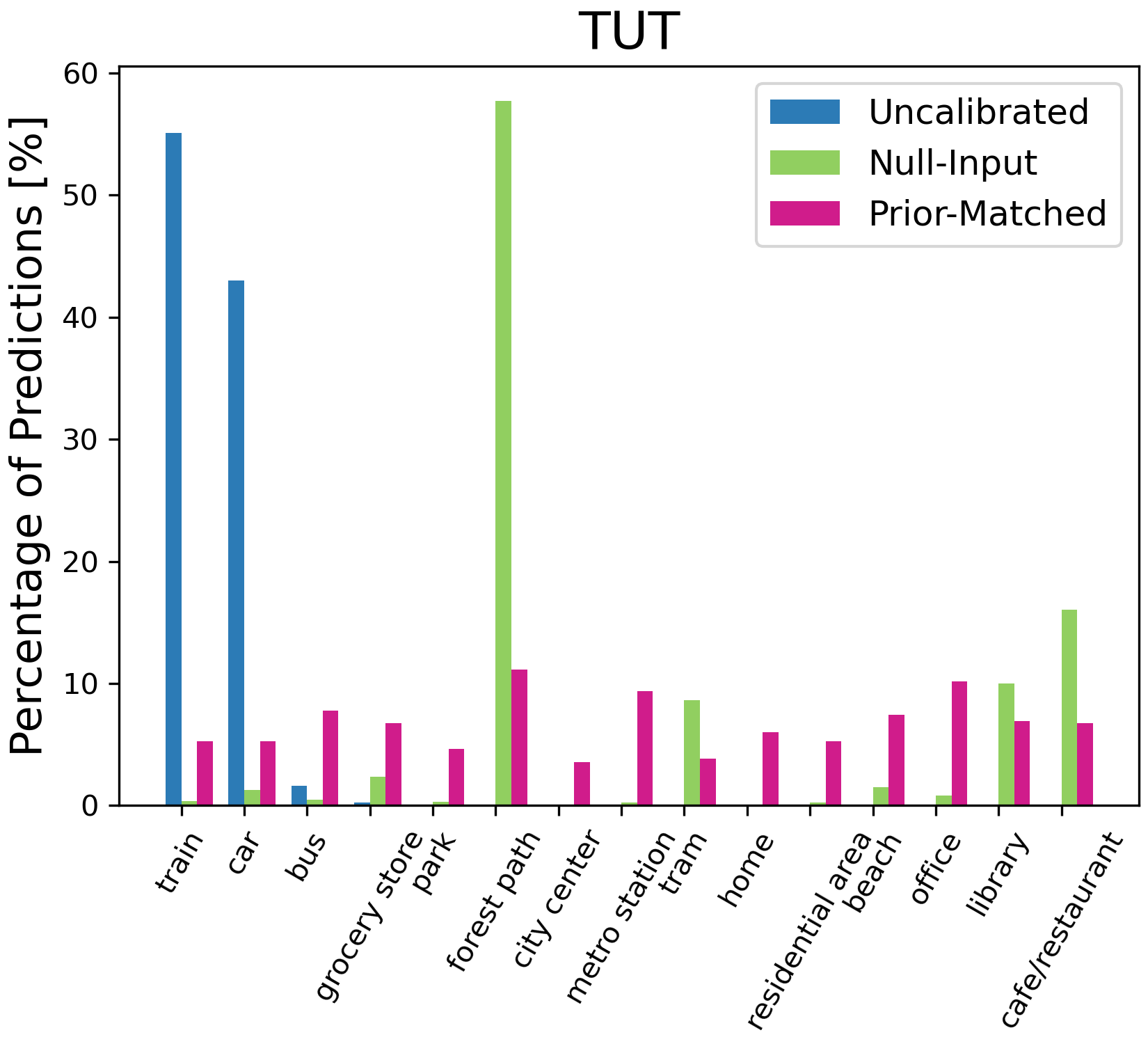}
    \includegraphics[width=0.24\columnwidth]{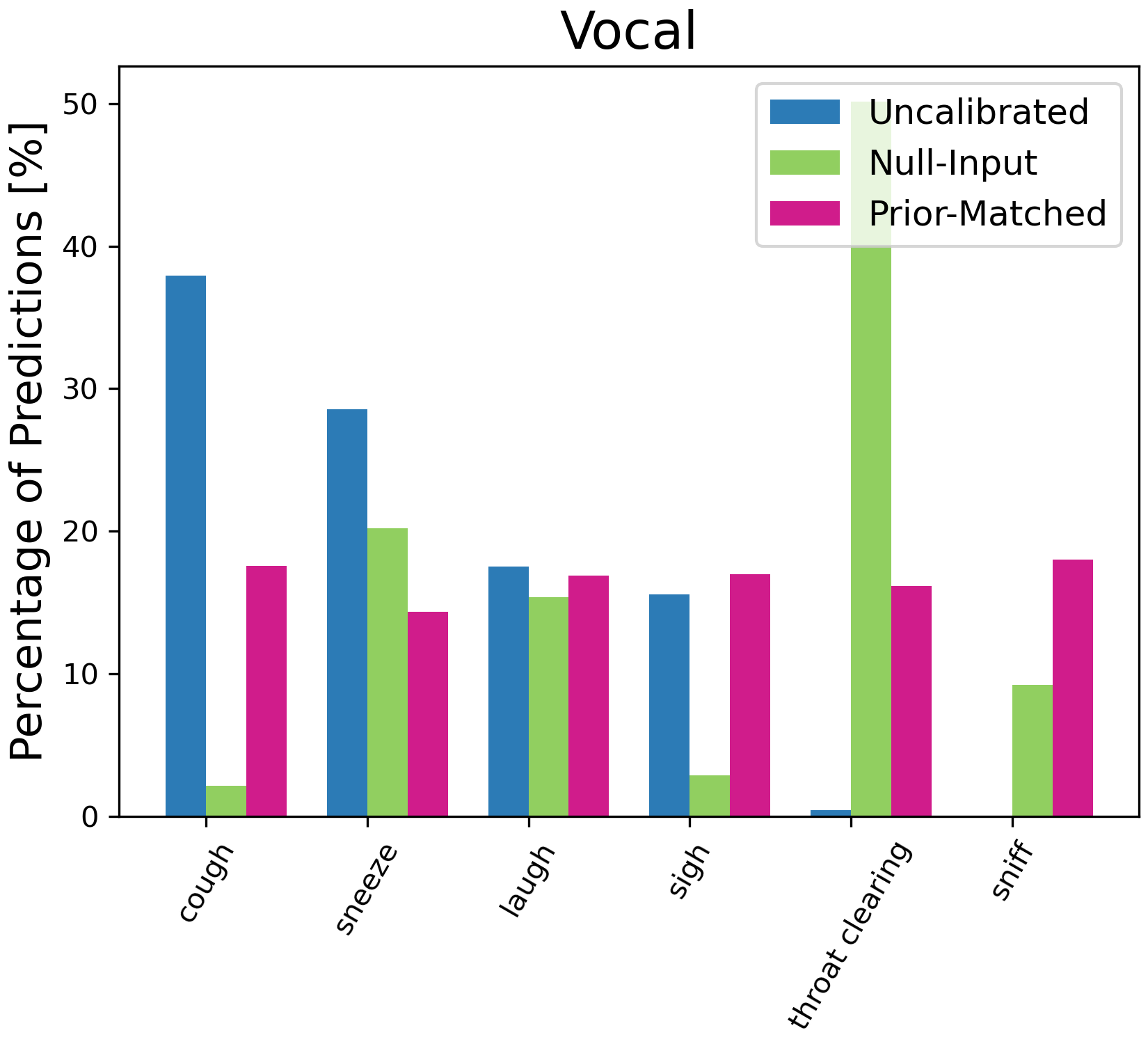}
    \includegraphics[width=0.24\columnwidth]{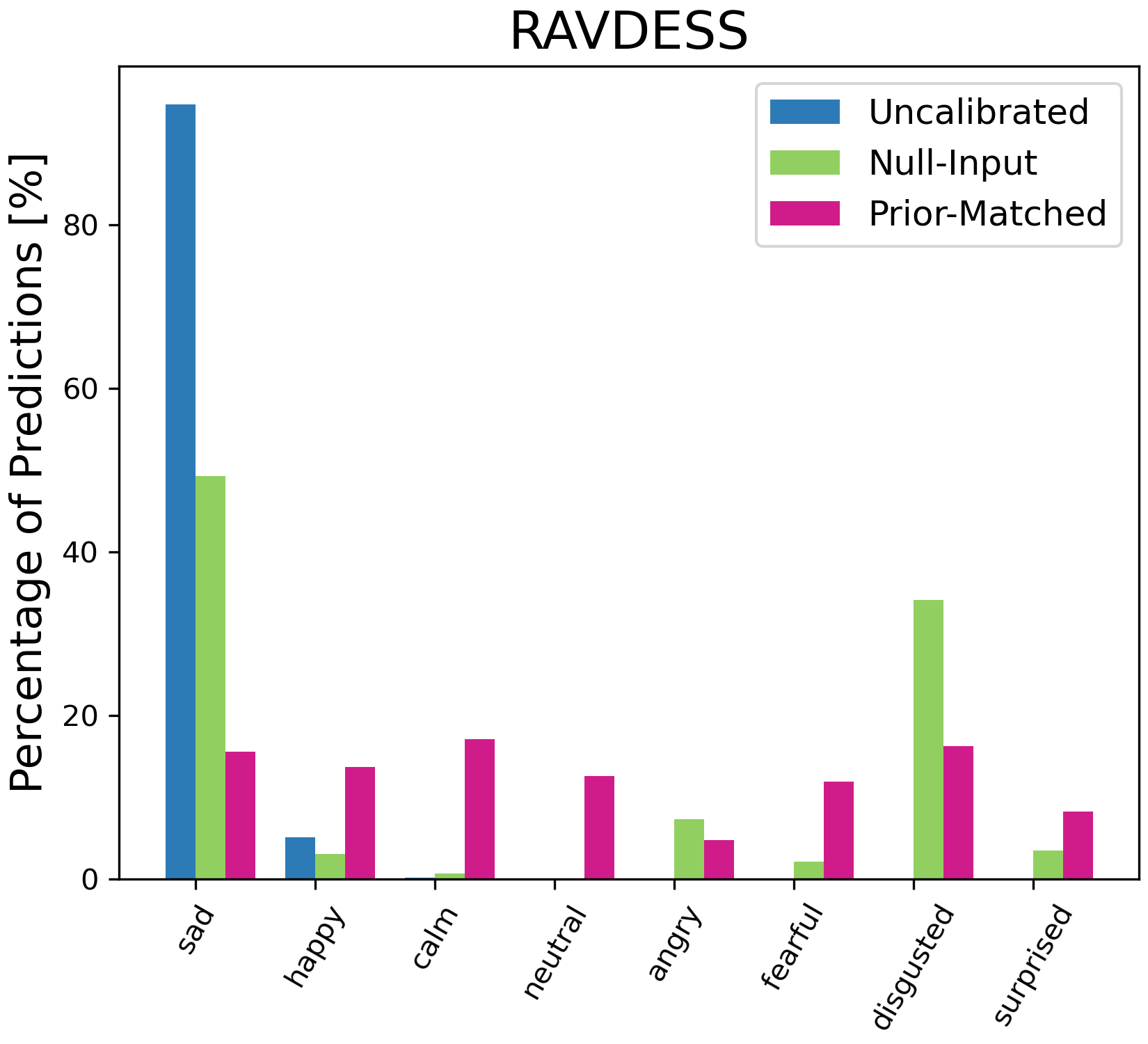}
    \includegraphics[width=0.24\columnwidth]{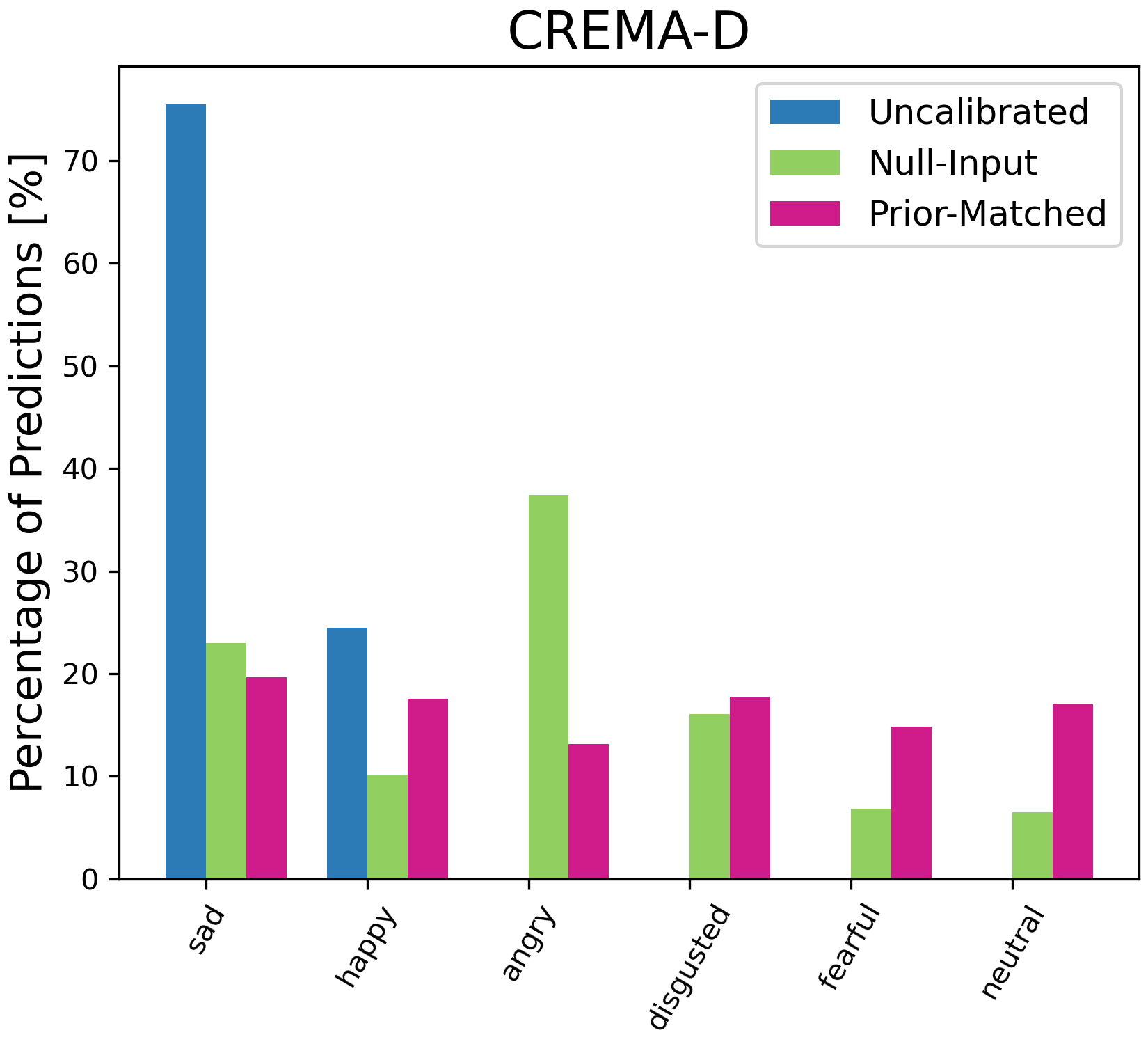}
    \includegraphics[width=0.24\columnwidth]{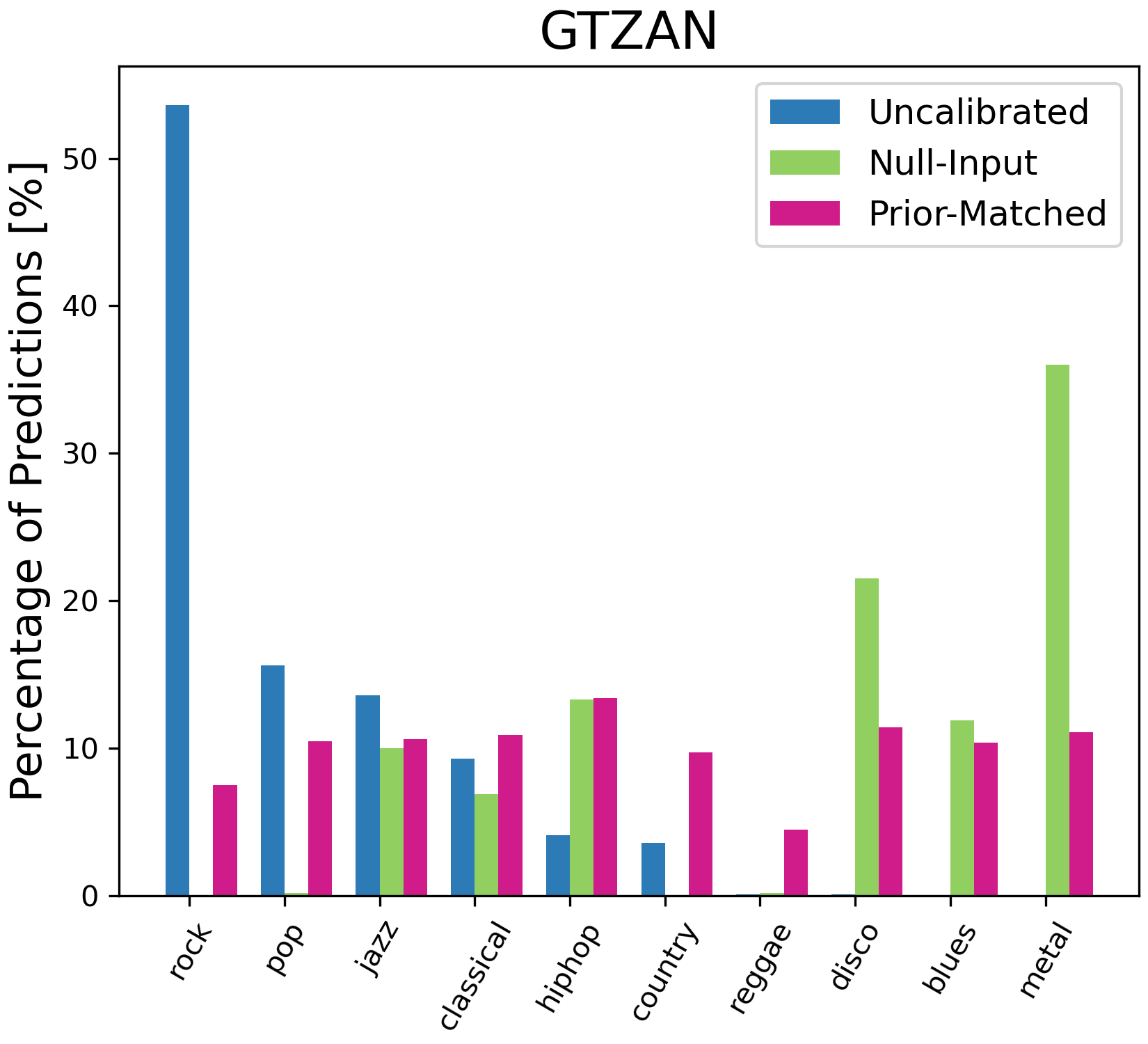}
    \includegraphics[width=0.24\columnwidth]{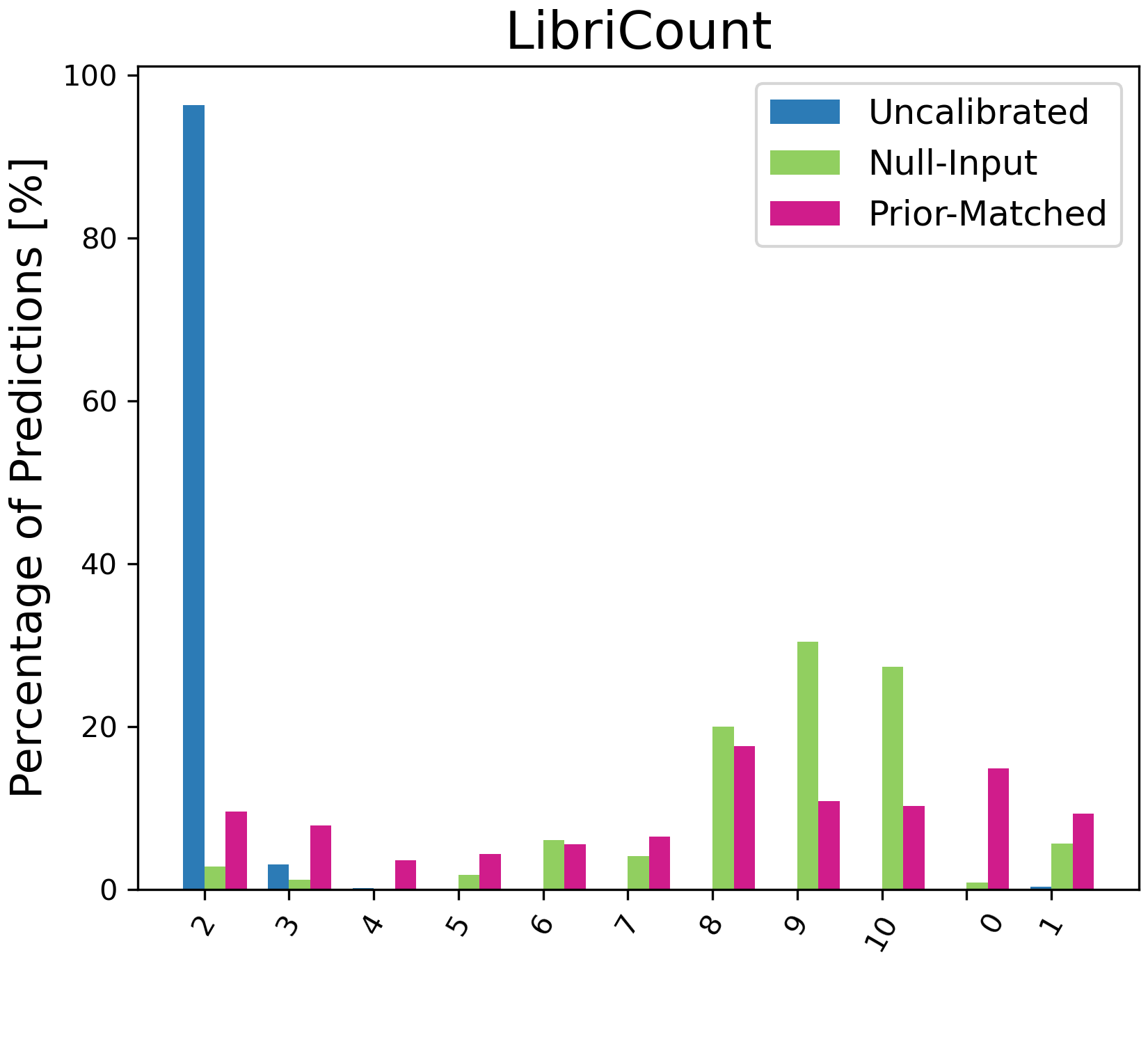}
    \caption{Percentage of model predictions for each class with different calibration methods. On ESC-50, we only plot the top 15 classes predicted by the uncalibrated results for illustration.}
    \label{fig:predictions}
\end{figure*}
\noindent Figure \ref{fig:predictions} shows the distribution of predicted classes for all the test samples on each dataset. For the uncalibrated results, the predictions are unevenly distributed among all the classes. Specifically, the system has a strong bias to predict words that are more likely to frequently appear in the pre-training data, such as \textit{`rain', `train', or `sad'}. Certain classes are never predicted due to the bias. This problem can be mitigated with null-input calibration. With prior matching, we can observe more evenly distributed predictions on the test samples.

\section{Robustness to Prompts}
\label{sec:appendix_prompts}

\begin{table*}[!htbp]
    \centering
    \fontsize{8}{10}\selectfont
    \begin{tabular}{l|c|c|c|c}
    \toprule
        Prompt & ESC50 & US8K & TUT & Vocal \\
        \midrule
        This is a sound of \textit{class\_label}. & 65.4 & 60.4 & 26.0 & 84.9 \\
        \midrule
        \textit{class\_label} & 48.6 & 54.8 & 15.7 & 60.1 \\
        \textit{(class\_label)} & 68.0 & 65.5 & 21.3 & 86.3 \\
        \textit{[class\_label]} & 64.3 & 64.2 & 16.1 & 85.9 \\
        \midrule
        Listen to the sound, it's called \textit{class\_label}. & 50.3 & 56.5 & 16.0 & 81.7 \\
        The noise you hear is from the category \textit{class\_label}. & 54.6 & 55.1 & 19.3 & 79.7\\
        This is what we call \textit{class\_label} sound. & 45.3 & 55.7 & 26.7 & 69.5 \\
        Identify this noise as \textit{class\_label}. & 46.6 & 52.8 & 13.6 & 81.6 \\
        This sound belongs to the group \textit{class\_label}. & 41.4 & 57.0 & 15.0 & 76.1 \\
        \midrule
        Ensemble of Prompts & 67.1 & 67.6 & 25.2 & 87.3 \\
    \bottomrule
    \end{tabular}
    \caption{Prompt sensitivity for Sound Event, Vocal Sound and Acoustic Scene Classification.}
\end{table*}

\begin{table*}[!htbp]
    \centering
    \fontsize{8}{10}\selectfont
    \begin{tabular}{l|c|c}
    \toprule
        Prompt & RAVDESS & CREMA-D \\
        \midrule
        The speaker is feeling \textit{class\_label}. & 41.7 & 28.8 \\
        \midrule
        \textit{class\_label} & 20.7 & 18.1 \\
        \textit{(class\_label)} & 33.1 & 35.3 \\
        \textit{[class\_label]} & 32.6 & 26.6 \\
        \midrule
        The person talking feels \textit{class\_label}. & 38.5 & 29.6 \\
        The speaker is experiencing \textit{class\_label} emotions. & 20.8 & 20.5 \\
        The person speaking is in a \textit{class\_label} mood. & 29.9 & 27.4 \\
        The speaker's emotion is \textit{class\_label}. & 33.6 & 25.1 \\
        The person talking is filled with \textit{class\_label} feelings. & 39.7 & 33.0 \\
        \midrule
        Ensemble of Prompts & 44.0 & 33.1 \\
    \bottomrule
    \end{tabular}
    \caption{Prompt sensitivity for Emotion Classification.}
\end{table*}

\begin{table}[ht]
\fontsize{7.5}{9}\selectfont
  \begin{minipage}[b]{0.47\linewidth}
    \centering
    \begin{tabular}{c|c}
        \toprule
        Prompt & GTZAN \\
        \midrule
        This is an audio of \textit{class\_label} music. & 60.9 \\
        \midrule
        \textit{class\_label} & 39.0 \\
        \textit{(class\_label)} & 54.6 \\
        \textit{[class\_label]} & 52.3 \\
        \midrule
        Listen to this, it's \textit{class\_label} music. & 48.5 \\
        This audio plays \textit{class\_label} music. & 38.8 \\
        The sound is from \textit{class\_label} music. & 49.4 \\
        What you're hearing is \textit{class\_label} music. & 58.7 \\
        This records \textit{class\_label} music. & 40.0 \\
        \midrule
        Ensemble of Prompts & 60.0 \\
        \bottomrule
    \end{tabular}
    \label{table1}
    \caption{Prompts for Music Genre Classification.}
  \end{minipage}%
  \begin{minipage}[b]{0.47\linewidth}
    \centering
    \begin{tabular}{c|c}
    \toprule
        Prompt & LibriCount \\
        \midrule
        In the audio, \textit{class\_label} people are speaking. & 17.3 \\
        \midrule
        \textit{class\_label} people speaking & 13.0 \\
        \textit{(class\_label people speaking)} & 15.3 \\
        \textit{[class\_label people speaking]} & 23.2 \\
        \midrule
        You can hear \textit{class\_label} people talking in the audio. & 9.2 \\
        The audio includes voices of people from \textit{class\_label}. & 14.6 \\
        In this recording, individuals from \textit{class\_label} are speaking. & 13.5\\
        The audio captures conversations of \textit{class\_label} individuals. & 11.6 \\
        The voices you're hearing are from \textit{class\_label} people. & 17.1 \\
        \midrule
        Ensemble of Prompts & 22.0 \\
    \bottomrule
    \end{tabular}
    \label{table2}
    \caption{Prompts for Speaker Counting.}
  \end{minipage}
\end{table}

\noindent The above tables show the performance of various decoder prompts for all the considered tasks. We observe that for some tasks, the natural language prompts are able to perform better than the class label-only prompt (TUT, RAVDESS, GTZAN), while for the other datasets, one may observe similar performance between our default prompts and class-only prompts.

\twocolumn
\section{Supervised Training Performance}
\label{sec:appendix_sup}

Two forms of efficient fine-tuning approaches are considered as supervised baselines; LoRA \cite{hu2021lora} and soft prompt tuning (SPT) \cite{lester2021power, ma2023adapting}. During training, the audio clip is provided to the model encoder and the model is trained to generate the corresponding class label in the decoder. For LoRA, we use a rank $r=8$ and only adapt the attention weights \cite{hu2021lora}. For SPT, we insert 20 learnable soft prompt vectors at the decoder input. This results in 940K $(0.06\%)$ and 25K $(0.002\%)$ learnable parameters for LoRA and SPT, respectively. During training, we use a batch size of 8, run 4000 training steps, use the AdamW optimizer with linear decay, and the learning rate is set to $1e^{-3}$ and $1e^{-1}$ for LoRA and SPT, respectively. Experiments are conducted on Whisper large-v2 for TUT and Vocal, which are the only of the considered tasks with available training data.

\begin{table}[!htbp]
    \centering
    \small
    \begin{tabular}{l|l|c|c}
    \toprule
        Method & Model & TUT & Vocal \\
    \midrule
        \multirow{3}*{Zero-shot} & Random & 6.7 & 16.7 \\
        & CLAP & 29.6 & 60.1 \\
        & Whisper & 26.0 & 84.9 \\
    \midrule
       \multirow{3}*{Supervised} 
        & CLAP & 74.6 & 97.9 \\
        & LoRA (Whisper)& 62.7 & 94.5 \\
        & SPT (Whisper) & 59.2 & 92.6 \\
    \bottomrule
    \end{tabular}
    \caption{Supervised training results on TUT and Vocal.}
    \label{tab:finetune_perf}
\end{table}

\noindent Table \ref{tab:finetune_perf} shows performance on TUT and Vocal, where as expected there remains a significant performance gap between the zero-shot and the supervised approaches. LoRA shows considerable performance improvements while being parameter efficient (and only learning 0.06\% of parameters). Supervised trained CLAP demonstrates better performance than Whisper, possibly as CLAP generates contextual embeddings that may be better suited for transferring to tasks, while Whisper is an ASR decoding system that typically isn't finetuned for downstream audio classification tasks.

\end{document}